\documentclass[preprint,5p,times,twocolumn,authoryear]{elsarticle}

\usepackage{amssymb}
\usepackage{amsmath}
\usepackage{amsthm}

\usepackage{lineno}

\usepackage{cite}
\usepackage{amsmath,amssymb,amsfonts}
\usepackage{algorithmic}
\usepackage{graphicx}
\usepackage{textcomp}
\usepackage{svg}
\usepackage{url}
\usepackage[citecolor=blue, urlcolor=blue, colorlinks=true]{hyperref}
\usepackage{booktabs}
\usepackage{diagbox}
\usepackage{multirow}
\usepackage{makecell}
\usepackage{balance}
\usepackage{enumitem}
\usepackage{cleveref}
\usepackage{caption}
\usepackage[most]{tcolorbox}
\usepackage{xurl}

\Crefname{figure}{Fig.}{Figs.}

\journal{Medical Image Analysis}

\begin{document}

\captionsetup[figure]{labelfont={bf},labelformat={default},labelsep=period,name={Fig.}}

\begin{frontmatter}



\title{SurgiATM: A Physics-Guided \textcolor{black}{Lightweight} Model for Deep Learning-Based Smoke Removal in Laparoscopic Surgery} 



\author[usyd]{Mingyu Sheng} 
\author[usyd]{Jianan Fan\corref{cor1}} 
\author[usyd]{Dongnan Liu} 
\author[sjt]{Guoyan Zheng} 
\author[hvd]{Ron Kikinis} 
\author[usyd]{Weidong Cai\corref{cor1}} 
\cortext[cor1]{Corresponding authors. \\ \textit{E-mail addresses}: \\ \url{jfan6480@uni.sydney.edu.au} (J. Fan), \url{tom.cai@sydney.edu.au} (W. Cai).}

\affiliation[usyd]{organization={School of Computer Science, The University of Sydney},
            city={Sydney},
            state={NSW},
            country={Australia}}
            
\affiliation[sjt]{organization={Institute of Medical Robotics, School of Biomedical Engineering, Shanghai Jiao Tong University},
            city={Shanghai},
            country={China}}
            
\affiliation[hvd]{organization={Department of Radiology, Brigham and Women's Hospital, and Harvard Medical School},
            city={Boston},
            state={MA},
            country={United States}}

\begin{abstract}
During laparoscopic surgery, smoke generated by tissue cauterization can significantly degrade the visual quality of endoscopic frames, increasing the risk of surgical errors and hindering both clinical decision-making and computer-assisted visual analysis. Consequently, removing surgical smoke is critical to ensuring patient safety and maintaining operative efficiency. 
In this study, we propose the Surgical Atmospheric Model (SurgiATM) for surgical smoke removal. SurgiATM statistically bridges a physics-based atmospheric model and data-driven deep learning models, combining the superior generalizability of the former with the high accuracy of the latter. \textcolor{black}{Furthermore, SurgiATM is designed as a lightweight module that can be easily integrated into existing surgical desmoking architectures with minimal modification, aiming to enhance their accuracy and stability.} 
The proposed method is derived via statistically optimizing a Mixture-of-Experts (MoE) model at the output end of arbitrary deep learning methods, with a Laplacian-like error distribution specifically leveraged to model surgical smoke. The output-stage MoE ensures minimal modification to the architecture of the original methods, while the Laplacian-like distribution characteristic of surgical smoke enables a lightweight reconstruction formulation with minimal parameters. 
Therefore, SurgiATM introduces only two hyperparameters and no additional trainable weights, preserving the original network architecture with minimal computational and modification overhead.
We conduct extensive experiments on three public surgical datasets with ten desmoking methods, involving multiple network architectures and covering diverse procedures, including cholecystectomy, partial nephrectomy, and diaphragm dissection. The results demonstrate that incorporating SurgiATM commonly reduces the restoration errors of existing models and relatively enhances their generalizability, without adding any trainable layers or weights. This highlights the convenience, low cost, effectiveness, and generalizability of the proposed method. The code for SurgiATM is released at \hyperlink{https://github.com/MingyuShengSMY/SurgiATM}{https://github.com/MingyuShengSMY/SurgiATM}.
\end{abstract}




\end{frontmatter}



\section{Introduction}
\label{Introduction}

Minimally invasive surgery (MIS) has noticeably benefited patients by reducing risk and enabling faster recovery \citep{Maier_Hein_2017_Surgical_data, Maier_Hein_2022_Surgical_data}. However, surgical smoke, commonly generated by high-energy instruments (e.g., electrosurgical units, laser surgical tools, and high-frequency ultrasonic scalpels), can severely impair endoscopic visibility, posing a notable challenge \citep{Duelmer_2026_2_}. 
The presence of dense smoke obstructs the surgeon's field of view, increases the risk of surgical errors, and reduces procedural efficiency.
Moreover, smoke-induced image degradation adversely affects vision-based downstream surgical tasks, such as 
phase recognition \citep{Yang_2024_Surgformer_Surgical, Zou_2025_Capturing_action, Wagner_2023_Comparative_validation, Yue_2023_Cascade_Multi, Guan_2024_Label_Guided, Zhang_2026_CSAP_Assist, Das_2025_PitVis_2023}, 
tool detection, 
semantic segmentation \citep{Sheng_2025_AMNCutter_Affinity, Sheng_2024_Revisiting_Surgical, Yue_2024_SurgicalSAM_Efficient, Sestini_2025_SAF_IS, Cartucho_2024_SurgT_challenge}, 
depth estimation \citep{Zou_2025_SSIFNet_Spatialtemporal, Tao_2023_SVT_SDE, Cheng_2022_Deep_Laparoscopic}, 
3D surface reconstruction \citep{Gong_2024_Self_Supervised, Maier_Hein_2014_Comparative_Validation, Rau_2024_SimCol3D_}, 
etc., 
which require both high frame quality and real-time processing speed 
\citep{Han_2025_Towards_Reliable, Long_2025_Surgical_embodied, Wang_2025_Learning_dissection, Xu_2026_Surgical_Action, Dou_2025_Artificial_intelligence, Rueckert_2026_Comparative_validation}.
At present, mechanical smoke evacuation or filtration is commonly achieved through hardware-based systems, which often require additional effort and impose extra workload during surgery, disrupting the surgical workflow and causing surgeon fatigue \citep{Manning_2018_Laparoscopic_lens, Xia_2025_In_Vivo}.
Therefore, ensuring a clear visual field in real time is critical \citep{Ulmer_2008_The_hazards, Carbajo_Rodrguez_2009_Surgical_smoke, Weld_2007_Analysis_of}. 

To this end, real-time image processing has gained increasing attention as a means of digitally removing smoke during intraoperative visualization in a more streamlined and automated way by leveraging computer vision and deep learning techniques. It offers a cost-effective, low-risk, and efficient alternative for surgical smoke removal \citep{Luo_2017_Vision_Based, Chen_2020_De_smokeGCN, Salazar_Colores_2020_Desmoking_Laparoscopy, Li_2024_Multi_frequency, Wang_2024_Desmoking_of}. 
However, traditional models often struggle with this task, as they were originally designed for outdoor hazy environments. Unlike natural hazy images, in vivo surgical scenes present unique challenges, including complex physics parameter estimation due to non-uniform distribution of smoke, spatially heterogeneous illumination caused by radial light attenuation, inaccurate color restoration on non-smoke objects (e.g., surgical instruments), severe information loss resulting from highly dense smoke obscuration, and temporal incoherence of smoke occurrence due to the intermittent use of energy devices \citep{Modrzejewski_2020_Light_modelling, Azagra_2023_Endomapper_dataset, Rodrguez_Puigvert_2023_LightDepth_Single, Stelzer_2024_Generation_and, Kumar_2023_Understanding_surgical, Tian_2015_Single_Image, Ramesh_2023_Dissecting_self}.
To address these limitations, deep learning-based approaches have been increasingly adopted for their powerful non-linear mapping capabilities and ease of deployment in laparoscopic systems \citep{Chen_2022_DesmokeNet_A, Li_2022_Endoscopy_image}. These methods show considerable promise for surgical desmoking, offering more consistent color restoration and improved visual clarity. Nevertheless, due to the scarcity of paired desmoking benchmarks, most deep learning-based models face the challenge of generalizability, as they can produce high-quality predictions on a dataset with synthetic smoke while suffering from performance degradation on another dataset with real smoke. This is an important concern for real-world clinical applications \citep{Wu_2025_Self_Supervised}. 

To overcome these limitations, we propose the \textbf{Surgi}cal \textbf{At}mospheric \textbf{M}odel (\textbf{SurgiATM}), a physics-guided model tailored to the laparoscopic desmoking task. SurgiATM can be seamlessly embedded into existing neural networks with minimal modifications to their original architectures, while enhancing both performance and generalizability. Notably, SurgiATM introduces \textbf{no additional trainable parameters}, indicating that the observed improvements arise from the integration of physical guidance rather than merely increasing model trainable weights and complexity.
Specifically, the main contributions of this study are:
\begin{itemize}
    \item \textcolor{black}{We propose SurgiATM, a parameter-efficient physics-guided model that improves performance and generalizability of existing surgical desmoking approaches, providing clearer desmoking results.} 
    
    \item The proposed model is derived via a Mixture-of-Experts (MoE) model at the output end, and thus SurgiATM can be employed as an output layer (or module), seamlessly integrated into state-of-the-art (SOTA) methods built upon various backbone architectures (e.g., U-Net, Swin Transformer).
    
    \item We develop SurgiATM based on statistical analysis and a physical model; thereby, it introduces zero trainable weights, while only two hyper-parameters are involved. The recommended configurations for the two hyper-parameters are analyzed in ablation studies.
    
    \item We validate SurgiATM on three public surgical datasets, demonstrating improvements in both performance and generalizability across multiple SOTA desmoking methods and various laparoscopic surgical scenes, thereby confirming its effectiveness, low efficiency overhead, stability, and generalizability for laparoscopic surgical smoke removal.
\end{itemize}

\section{Related Work}
\label{RelatedWork}


    \subsection{Physics-Based Methods}
        The Atmospheric Scattering Model (ASM) is a widely used physical model that describes the formation of hazy natural images by modeling the absorption, scattering, and reflection of light by objects and particles in the atmospheric medium \citep{Narasimhan_2002_Vision_and, Narasimhan_2003_Contrast_restoration}. 
        Based on the ASM, clear images can be restored from degraded observations by estimating the required physical parameters, such as medium transmission, atmospheric light, and scene depth, which has motivated extensive research in the field. 
        In early studies, the Gaussian distribution and Markov Random Field (MRF) model were classic and well-known techniques for estimating the parameters \citep{Fattal_2008_Single_image, Tan_2008_Visibility_in}. 
        Then, \citet{Kaiming_2011_Single_Image} proposed the Dark Channel Prior (DCP) model for natural image dehazing, in which the medium transmission and global atmospheric light were statistically estimated according to their physical properties. 
        The DCP assumes that haze in an image is uniformly distributed and that, in hazeless images, at least one color channel in most non-sky patches or pixels has very low intensity. In contrast, haze tends to raise pixel intensities across all channels, causing the dark channel to become brighter and the image to appear more whitish. 
        Since 2012, the field of surgical desmoking has gained increasing attention, with many valuable studies published over the past decade.
        \citet{Gu_2015_Virtual_Digital} developed a digital defogging system that uses the DCP to enhance image quality in laparoscopic surgery. 
        \citet{Kotwal_2016_Joint_desmoking} combined the desmoking and denoising tasks of laparoscopic images and transformed them into a Bayesian inference problem by proposing a novel probabilistic graphical model.
        \citet{Baid_2017_Joint_desmoking} extends \citep{Kotwal_2016_Joint_desmoking} by introducing additional image priors of color and texture derived from sparse dictionary models.
        In the \citet{Tchaka_2017_Chromaticity_based} study, histogram equalization is combined with an adaptive DCP to remove surgical smoke. 
        \citet{Luo_2017_Vision_Based} rearranged the original atmospheric model by introducing an atmospheric veil to represent the transmission map and then solved the parameter estimation problem via the Poisson equation in the frequency domain. 
        \citet{Wang_2018_Variational_based} first developed an energy function to optimize and estimate a smoke veil to recover smokeless images based on \citet{Tian_2015_Single_Image}. 
        Combining the hardware camera and the software algorithm, \citet{Azam_2022_Smoke_removal} proposed a method based on multiple-exposure fusion, which enhances local contrast and color saturation, thereby improving image quality.
        These works offer valuable insights into the feasibility of physics-based approaches for surgical desmoking, encouraging further research.
        
    \subsection{Deep Learning-Based Methods}
        To achieve higher restoration quality, the powerful fitting and generative capabilities of Deep Neural Networks (DNNs) offer an alternative solution \citep{Cai_2016_DehazeNet_An, Li_2017_AOD_Net, Dong_2020_Multi_Scale, Qin_2020_FFA_Net, Bolkar_2018_Deep_Smoke, Sidorov_2020_Generative_Smoke}. 
        Current deep learning-based desmoking methods can be broadly categorized into two frameworks: non-adversarial learning and adversarial learning.

        
        \subsubsection{Non-adversarial Learning Methods} 
        This framework indicates that the model is tasked with learning a mapping from smoky or hazy images to their corresponding clean versions. 
        \citet{Chen_2018_Unsupervised_Learning} trained a U-Net model \citep{Ronneberger_2015_U_Net} on laparoscopic videos with synthetic smoke using Blender\footnote{https://www.blender.org/}.
        \citet{Bolkar_2018_Deep_Smoke} attempted to address surgical smoke removal via AODNet \citep{Li_2017_AOD_Net}, a Convolutional Neural Network (CNN) designed to estimate global light illuminance and the transmission map simultaneously.
        Similar to \citet{Chen_2018_Unsupervised_Learning}, \citet{Bolkar_2018_Deep_Smoke} trained the model on synthetically generated smoky in vivo frames by blending Perlin noise with clean frames.
        Furthermore, instead of training on synthetic smoky frames, \citet{Ma_2021_A_Smoke} attempted to use adjacent frames, along with the presence or absence of smoke, as training inputs and corresponding ground truth. 
        Similarly, in the most recent study, \citet{Wu_2025_Self_Supervised} implemented a real-world self-supervised video desmoking approach by capturing the informative features from frames recorded prior to the activation of high-energy surgical instruments.
        Based on the above training data strategy, considerable efforts have been made to improve performance further.
        \citet{Wang_2019_Multiscale_deep} designed a U-Net-like network and proposed Laplacian image pyramid decomposition to capture multi-scale features. 
        \citet{Chen_2020_De_smokeGCN} presented a novel generative-collaborative learning framework (De-smokeGCN), dividing the desmoking into two sub-tasks: smoke detection and smoke removal. A smoke detector predicted the smoke mask, which was leveraged as a prior for the smoke removal network and motivated several subsequent studies \citep{Hong_2023_MARS_GAN, Zhou_2024_Synchronizing_Detection}. 
        \citet{Lin_2021_A_desmoking} improved the decoder part of the U-Net by appending a convolutional block attention module to generate a guidance mask for feature maps. 
        \citet{Kanakatte_2021_Surgical_Smoke} introduced a workflow to simultaneously remove smoke and restore colors, using U-Net as the backbone. 
        \citet{Ma_2024_Laparoscopic_Video} leveraged deformable convolution incorporated with a mutual attention mechanism to model temporal features and combined Local Binary Patterns with input frames as a texture prior.
        \citet{Wang_2024_Smoke_veil} developed a method consisting of three U-Nets to produce distinct outputs, each serving as an element in the energy function derived from their previous study \citep{Wang_2018_Variational_based}.
        \citet{Li_2024_Multi_frequency} developed a surgical desmoking method based on the Diffusion Model \citep{Ho_2020_Denoising_Diffusion, Jiaming_2021_Denoising_Diffusion}, incorporating a multi-level frequency analysis module to integrate features across different frequency bands.
        \textcolor{black}{Similarly, \citet{Chen_2026_EndoIR_Degradation} proposed EndoIR, a diffusion-based model featuring task-adaptive embeddings to handle multiple tasks, including desmoking, blood removal, and low-light image enhancement.}
        \citet{Wu_2025_Learning_Multi} further combined the information in frequency and temporal domains and developed a multi-scale Gaussian window constraint to capture inter-frame features with fine structural details preserved.
        In addition to convolutional network architectures, the Swin Transformer (SwinT) \citep{Liu_2021_Swin_Transformer} has recently gained increasing attention in this field because of its hierarchical representation and efficient local-to-global attention mechanism.
        \citet{Wang_2023_Surgical_smoke} proposed a network that utilized several convolutional layers to extract low-level features and then used SwinT to analyze deep and global information from smoky frames.
        \citet{Wang_2024_Desmoking_of} developed a U-shaped transformer model based on U-Net and SwinT to enhance feature representation by combining local detail extraction with global context modeling.
        

    \begin{figure*}[t]
    \centering
    \tcbox[colframe=red, colback=white, boxrule=0.25mm, size=tight]{\includegraphics[width=1.0\linewidth]{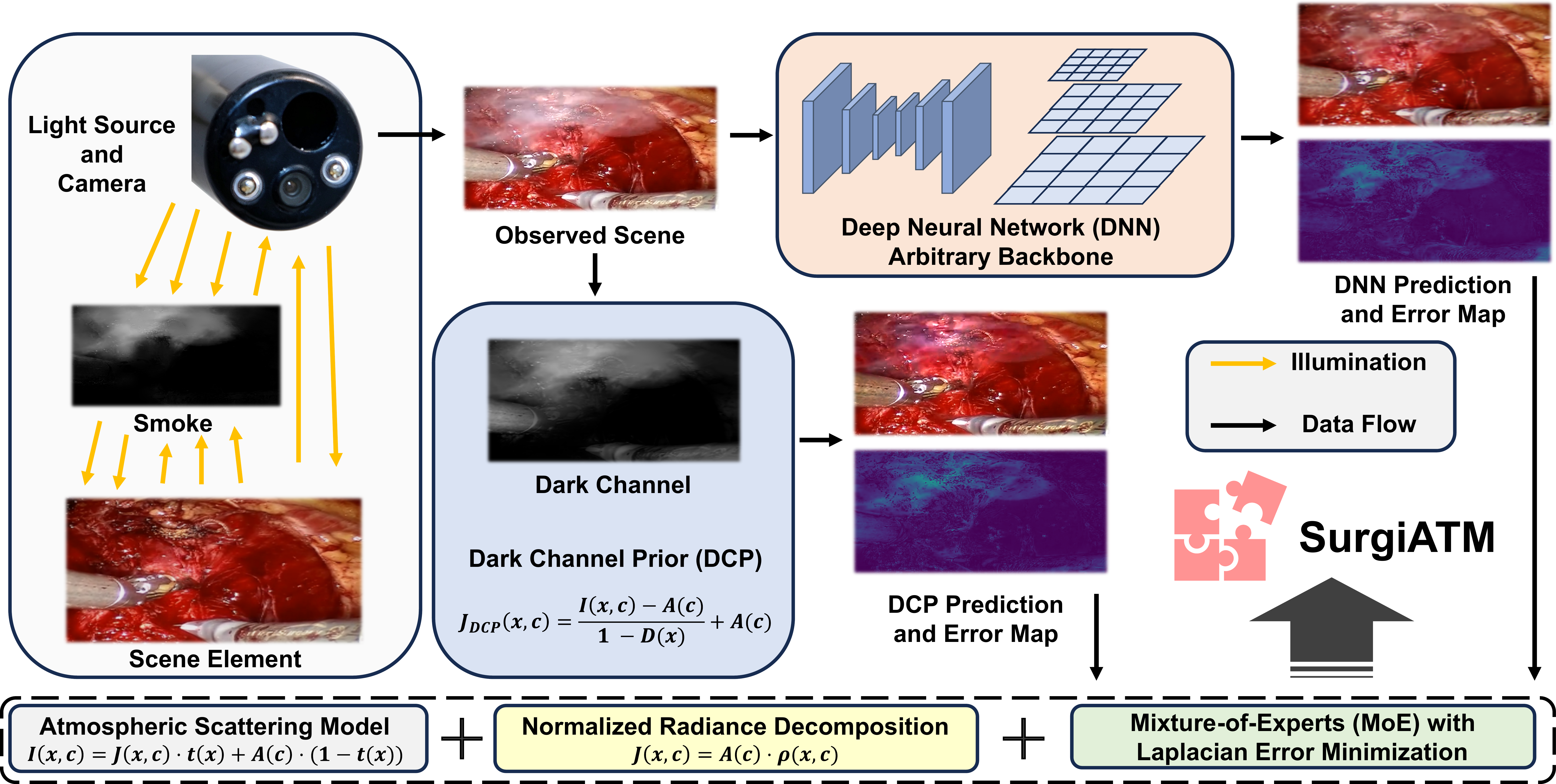}}
    \caption{The proposed SurgiATM is derived from the Atmospheric Scattering Model, Normalized Radiance Decomposition, Mixture-of-Experts (MoE), and Laplacian Error Minimization. We collect prediction results and errors from a lot of SOTA methods with diverse backbone architectures. Then the Laplacian error distribution is observed from surgical desmoking, which is different from natural fog. Compared to physics-based methods (e.g., DCP), data-driven methods usually demonstrate higher accuracy, while the former show strong stability. To complement their strengths, SurgiATM is developed by optimizing this system.}
    \label{Fig_Developing_Flow}
    \end{figure*}
        
        \subsubsection{Adversarial Learning Methods} 
        Apart from conventional image-to-image translation, converting laparoscopic images from the smoky domain to the smokeless domain offers an alternative paradigm for addressing the surgical desmoking problem. By regarding desmoking as an image style transfer task, the Generative Adversarial Network (GAN) \citep{Goodfellow_2020_Generative_adversarial} and its variants have emerged as effective solutions.
        \citet{Sidorov_2020_Generative_Smoke} modeled a mapping between smoky and smoke-free domains using a conventional GAN as the backbone and leveraging the perceptual image quality score as a loss function to improve the result quality.
        \citet{Salazar_Colores_2020_Desmoking_Laparoscopy} developed a method based on a Conditional Generative Adversarial Network (cGAN) \citep{Mirza_2014_Conditional_generative} and used an embedded dark channel as a prior.
        Compared to the above GANs, the Cycle-Consistent Adversarial Network (CycleGAN) \citep{Zhu_2017_Unpaired_Image} is well-suited for image style transfer tasks and is therefore more commonly used for surgical smoke removal \citep{Su_2023_Multi_stages, Wang_2025_Desmoke_VCU, Islam_2024_LVQE_Laparoscopic}.
        \citet{Vishal_2019_Guided_Unsupervised} enhanced CycleGAN by incorporating multi-scale feature extraction and introducing an upsampling loss to improve the contrast of the desmoked frames. In their subsequent studies, they further included structure-consistency loss and designed a refinement module to improve restoration quality \citep{Vishal_2020_Unsupervised_Desmoking, Venkatesh_2020_Unsupervised_smoke}. 
        \citet{Pan_2022_DeSmoke_LAP} further designed internal-channel and dark-channel loss functions based on the characteristics of smoky pixels to improve model performance.
        Similar to \citet{Chen_2020_De_smokeGCN}, \citet{Zhou_2024_Synchronizing_Detection} added an extra detection network to the CycleGAN architecture to estimate the smoke mask as a prior for subsequent desmoking.
        \citet{Hong_2023_MARS_GAN} adopted a multilevel strategy to adaptively learn non-homogeneous smoke features, using the predicted smoke distribution as a prior.


        To date, existing studies have made significant contributions; however, some limitations remain, as the feasibility and practicality of physics-based methods are primarily restricted by accuracy, while those of deep learning-based methods are mainly hindered by robustness and generalizability. In this paper, we conduct a statistical analysis of their respective advantages and subsequently derive a \textcolor{black}{lightweight and compatible} module to augment the existing end-to-end deep learning desmoking methods, further enhancing their performance and generalizability across different benchmarks.

    \begin{table}[h]
    \centering
    \caption{Primary Notations.}
    \begin{tabular}{l l}
        \toprule
        \textbf{Variable} \; \; \; & \textbf{Definition} \\
        \midrule

        $ I $ & Smoky image  \\
        $ J $ & Smoke-free image  \\
        $ J_\text{DCP}, J_\text{DNN} $ & $J$ predicted by DCP or DNN  \\
        $ t $ & Medium transmission  \\
        $ A $ & Global atmospheric light  \\
        $ \rho $ & Normalized radiance  \\
        $ D $ & Dark channel  \\
        $ \mathcal{D} $ & Denormalized dark channel  \\

        $ W $ & Expert-weights for MoE \\
        $ \varepsilon $ & Error term  \\
        $ \mu, b $ & Laplacian parameters  \\

        $ \eta $ & Smoothing factor  \\
        $ z $ & Window size  \\


        \bottomrule
    \end{tabular}
    \label{Tab_Notation}
    \end{table}

\section{Methods}

\begin{figure*}[t]
    \centering
    \includegraphics[width=1\linewidth]{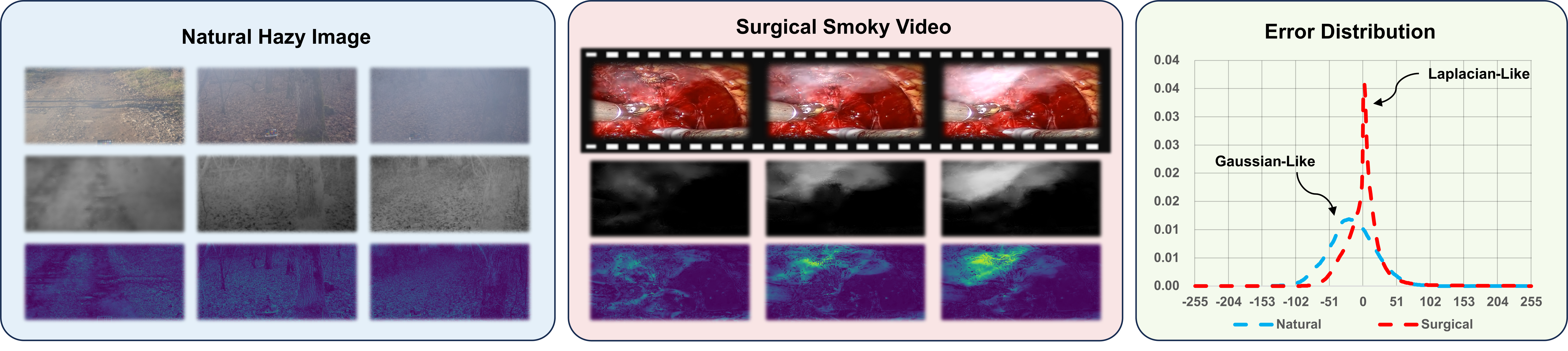}
    \caption{The left images are from a natural dehazing dataset, O-HAZE \citep{Ancuti_2018_O_HAZE}; the middle group is from a real-world surgical desmoking benchmark, VASST-desmoke \citep{Xia_2025_In_Vivo}; and the line chart (right) shows the difference in error distributions between surgical desmoking and natural dehazing, computed from the entire dataset. The $\text{1}^\text{st}$ row displays the hazy or smoky images; the $\text{2}^\text{nd}$ row shows the smoke or haze mask estimated from the ground truth; and the $\text{3}^\text{rd}$ row presents the error magnitude map of the DCP prediction.}
    \label{Fig_Natural_MIS_Error_Comparison}
    \end{figure*}

    The proposed method is motivated by an intriguing observation in laparoscopic surgical video smoke removal: predictions from most desmoking methods yield a Laplacian-like error distribution (see \Cref{Fig_Natural_MIS_Error_Comparison}). Building upon this insight, we develop a new restoration formula that minimizes prediction errors and enhances model robustness by revisiting classic physical models, analyzing the predictions of deep learning-based approaches, leveraging an MoE model, and optimizing from the perspective of probability distributions (see \Cref{Fig_Developing_Flow}). SurgiATM can be seamlessly embedded at the output stage of deep learning methods, where it predicts a normalized radiance used for reconstruction. Integration with general DNN baseline methods is demonstrated in \Cref{Fig_Plug_in_baseline}. The primary variables are summarized in \Cref{Tab_Notation}.
    
    \subsection{Background}
        \subsubsection{Physics-Based Modeling}
            The atmospheric scattering model \citep{Narasimhan_2002_Vision_and, Narasimhan_2003_Contrast_restoration, Fattal_2008_Single_image, Tan_2008_Visibility_in} is typically formulated as: 
            \begin{flalign}
                \label{Equ_Method_BG_ATM}
                &\ I(x, c) = J(x, c) \cdot t(x) + A(c) \cdot ( 1 - t(x) ), &
            \end{flalign}
            where $c$ denotes a color channel; $x$ represents the spatial location of a pixel; $I \in \mathbb{R}^{H \times W \times C}$ is the observed image, where $C$, $W$, and $H$ are the number of color channels, width and height, respectively; $J$ indicates the corresponding haze-free image; $A \in \mathbb{R}^{1 \times 1 \times C}$ denotes the global atmospheric light; and $t \in \mathbb{R}^{H \times W \times 1}$ is the medium transmission.
            In particular, $t$ and $J$ can be further formulated as: 
            \begin{flalign}
                \label{Equ_Method_BG_ATM_t}
                &\ t(x) = e^{-\beta(\lambda) \cdot d(x)}, &
            \end{flalign}
            \begin{flalign}
                \label{Equ_Method_BG_ATM_J}
                &\ J(x, c) = A(c) \cdot \rho(x, c), &
            \end{flalign}
            where $\beta(\lambda) \in \mathbb{R}$ is the scattering coefficient for a specific wavelength $\lambda$ of light; $d \in \mathbb{R}^{H \times W \times 1}$ represents the scene depth; and $\rho \in \mathbb{R}^{H \times W \times C}$ is the normalized radiance of a scene point, which is mainly determined by reflectance of object surfaces \citep{Narasimhan_2003_Contrast_restoration, Tan_2008_Visibility_in}.
            
    
            \citet{Kaiming_2011_Single_Image} proposed the Dark Channel Prior (DCP), which estimates the transmission map $t$ using the dark channel:
            \begin{flalign}
            &\ \begin{aligned}
                \label{Equ_Method_BG_DCP}
                t(x) 
                &= 1 - D(x) \\
                &= 1 - \underset{c \in C}{\mathrm{min}} \left( \underset{u \in \Omega(x; z)}{\mathrm{min}} \left( \frac{I(u, c)}{A(c)} \right) \right),
            \end{aligned}&
            \end{flalign}
            where $D$ denotes the dark channel and $\Omega(x)$ represents a square window of size $z$ centered at location $x$. Accordingly, the restoration formula is given by:
            \begin{flalign}
                \label{Equ_Method_BG_DCP_RES}
                &\ J_\mathrm{DCP}(x, c) = \frac{I(x, c) - A(c)}{ 1 - D(x) } + A(c),&
            \end{flalign}
            where $J_\mathrm{DCP}$ indicates that the image $J$ is estimated using DCP; $1 - D(x)$ is empirically clipped by a lower bound $t_0$ in practical implementations to prevent division by zero, while this application-based adjustment is omitted for simplicity and clarity of the subsequent mathematical derivation.
            
    
        \subsubsection{Deep Learning-Based Modeling}
        \label{Sec_Method_DL}
            Clearly, in pixels where the smoke density is high and white or gray surgical instruments are present (i.e., $D \rightarrow 1$), the conventional physics-based model Eq.~\eqref{Equ_Method_BG_DCP_RES} may fail. Moreover, radial light attenuation can violate the assumption that $A \in \mathbb{R}^{1 \times 1 \times C}$; under such conditions, $A$ is more appropriately represented as $A \in \mathbb{R}^{H \times W \times C}$. Adhering to the original assumption may introduce more errors, whereas adopting the latter formulation significantly increases the degrees of freedom of the solution space. Therefore, Deep Neural Network (DNN), a powerful technique for non-linear mapping, is widely employed for smoke removal. Its formulation and training objective are typically expressed as:
            \begin{flalign}
                \label{Equ_Method_BG_DNN_RES}
                &\ J_\mathrm{DNN}(x, c) = \Phi_\theta \left( I(x, c), \tilde{Z} \right), & \\
                \label{Equ_Method_BG_DNN_GENERAL_LOSS}
                &\ \theta^* = \underset{\theta}{\arg \min} \; \; \mathcal{L} (J, J_\mathrm{DNN}; \mathcal{X}, \theta), &
            \end{flalign}
            where $\Phi$ indicates a non-linear mapping parameterized by trainable weights $\theta$; $\theta^*$ represents the optimized weights obtained through training on the dataset $\mathcal{X}$; $\tilde{Z}$ and $\mathcal{L}$ denote priors and the loss function, respectively, as determined by the specific method design.
            Nevertheless, its generalizability may be limited when deployed in unseen surgical scenarios.
            
    
    \subsection{Mixture-of-Experts and Optimization}   
    \label{Sec_MoE_Opt}
        Inspired by the Mixture-of-Experts (MoE), an effective mechanism for complementary advantages \citep{Jacobs_1991_Adaptive_Mixtures, Ding_2025_DenseFormer_MoE}, we begin with an ideal gating model: 
        \begin{flalign}
        &\ \begin{aligned}
            \label{Equ_Method_EMM_MoE_1}
            J_\mathrm{MoE}(x, c) &= \tilde{W}(x) \cdot J_\mathrm{DCP}(x, c) \\
            & \quad + (1 - \tilde{W}(x)) \cdot J_\mathrm{DNN}(x, c),
        \end{aligned} &
        \end{flalign}
        where $J_\mathrm{MoE}$ is the restored result of the gating model, and $\tilde{W} \in \{0, 1\}^{H \times W \times 1}$ is a pixel-wise indicator specifying which method is activated for smoke removal. Specifically, $\tilde{W}(x) = 1$ if $J_\mathrm{DCP}(x, c)$ is deemed more reliable than $J_\mathrm{DNN}(x, c)$; otherwise, $\tilde{W}(x) = 0$. Optimizing $\tilde{W}$ facilitates the most accurate prediction. For simplification and tractability, $\tilde{W}$ is relaxed to a continuous domain, yielding $W \in [0, 1]^{H \times W \times 1}$, and the restorations are formulated using additive error models:
       \begin{flalign}
            \label{Equ_Method_EMM_DCP_JJE}
            &\ J_\mathrm{DCP}(x, c) = J(x, c) + \varepsilon_\mathrm{DCP}(x, c), & \\
            \label{Equ_Method_EMM_DNN_JJE}
            &\ J_\mathrm{DNN}(x, c) = J(x, c) + \varepsilon_\mathrm{DNN}(x, c), &
        \end{flalign}
        where the predicted image is represented as the ground truth augmented by an independent and identically distributed (i.i.d.) error term $\varepsilon \in \mathbb{R}^{H \times W \times C}$. The error term is commonly assumed to follow a zero-mean Gaussian distribution with variance $\sigma^2$, denoted as $\varepsilon(x, c) \overset{\mathrm{i.i.d.}}{\sim} \mathcal{N}(\mu=0, \sigma^2)$, which is a classic assumption in statistical learning.
        Then, we have: 
        \begin{flalign}
            \label{Equ_Method_EMM_MoE_2}
            &\ J_\mathrm{MoE}(x, c) = J(x, c) + \varepsilon_\mathrm{MoE}(x, c), & \\
        &\ \begin{aligned}
            \label{Equ_Method_EMM_MoE_ERR}
           \varepsilon_\mathrm{MoE}(x, c) = W(x) \cdot \varepsilon_\mathrm{DCP}(x, c)  \\
            \quad + (1 - W(x)) \cdot \varepsilon_\mathrm{DNN}(x, c). 
        \end{aligned} &
        \end{flalign}
        Clearly, optimizing $J_\mathrm{MoE}$ is equivalent to minimizing the error term $\varepsilon_\mathrm{MoE}$: 
        \begin{flalign}
        &\ \begin{aligned}
            \label{Equ_Method_EMM_MoE_ERR_OPT}
            \min_{W(x)} \; \; & \mathbb{E} [ \varepsilon_{\mathrm{MoE}}^2 ], \quad
            s.t. \; \; 0 \leq W(x) \leq 1.
        \end{aligned} &
        \end{flalign}
        
        \begin{table}[t]
            \caption{Comparison of Prediction Error Distributions with Gaussian and Laplacian Distributions via Jensen-Shannon (JS) Divergence in VASST-desmoke with real-world smoke and Cholec80 with synthetic smoke. \textcolor{black}{\textbf{Bold} means the error distribution of each method is closer to a Gaussian or Laplacian distribution.}}
            \label{Table_VASST_Cholec_JS_Gau_Lap_Comp}
            \centering
            \resizebox{0.5\textwidth}{!}{
            \begin{tabular}[width=1.0\linewidth]{l c c c c }
                \toprule
                \multirow{2}{*}{\textbf{Method}} & \multicolumn{2}{c}{\textbf{VASST-desmoke}} & \multicolumn{2}{c}{\textbf{Cholec80}} \\
                
                ~ & \textbf{Gaussian} & \textbf{Laplacian} & \textbf{Gaussian} & \textbf{Laplacian} \\
                
                \midrule
                
                DCP \citep{Kaiming_2011_Single_Image} & 0.044989 & \textbf{0.013664} & \textbf{0.002782} & 0.010886 \\
                
                AODNet \citep{Li_2017_AOD_Net} & 0.077348 & \textbf{0.056151} & \textbf{0.006405} & 0.012263 \\
                
                CGAN-DC \citep{Salazar_Colores_2020_Desmoking_Laparoscopy}& \textbf{0.033480} & 0.033894 & 0.007939 & \textbf{0.005278} \\
                
                De-smokeGCN \citep{Chen_2020_De_smokeGCN} & 0.027138 & \textbf{0.001988} & 0.007484 & \textbf{0.005802} \\
                
                GCANet \citep{Chen_2019_Gated_Context} & 0.009448 & \textbf{0.003621} & \textbf{0.003164} & 0.008474 \\
                
                LGUTransformer \citep{Wang_2024_Desmoking_of} & 0.010483 & \textbf{0.003129} & 0.016931 & \textbf{0.003623} \\
                
                MARS-GAN \citep{Hong_2023_MARS_GAN} & \textbf{0.024441} & 0.064570 & 0.009878 & \textbf{0.005369} \\
                
                MSBDN \citep{Dong_2020_Multi_Scale} & 0.020062 & \textbf{0.007690} & 0.005995 & \textbf{0.005536} \\
                
                RSTN \citep{Wang_2023_Surgical_smoke} & 0.026959 & \textbf{0.006900} & 0.012523 & \textbf{0.004193} \\
                
                SSIM-PAN \citep{Sidorov_2020_Generative_Smoke} & 0.018315 & \textbf{0.011999} & \textbf{0.005900} & 0.006517 \\
                
                SVPNet \citep{Wang_2024_Smoke_veil} & 0.022240 & \textbf{0.000675} & 0.007876 &\textbf{ 0.006492} \\
                
                \midrule

                Avg. & 0.028628 & \textbf{0.013795} & 0.008734 & \textbf{0.006734} \\
                
                \bottomrule
            \end{tabular}
            }
        \end{table}
        \begin{figure*}[t]
        \centering
        \includegraphics[width=1\linewidth]{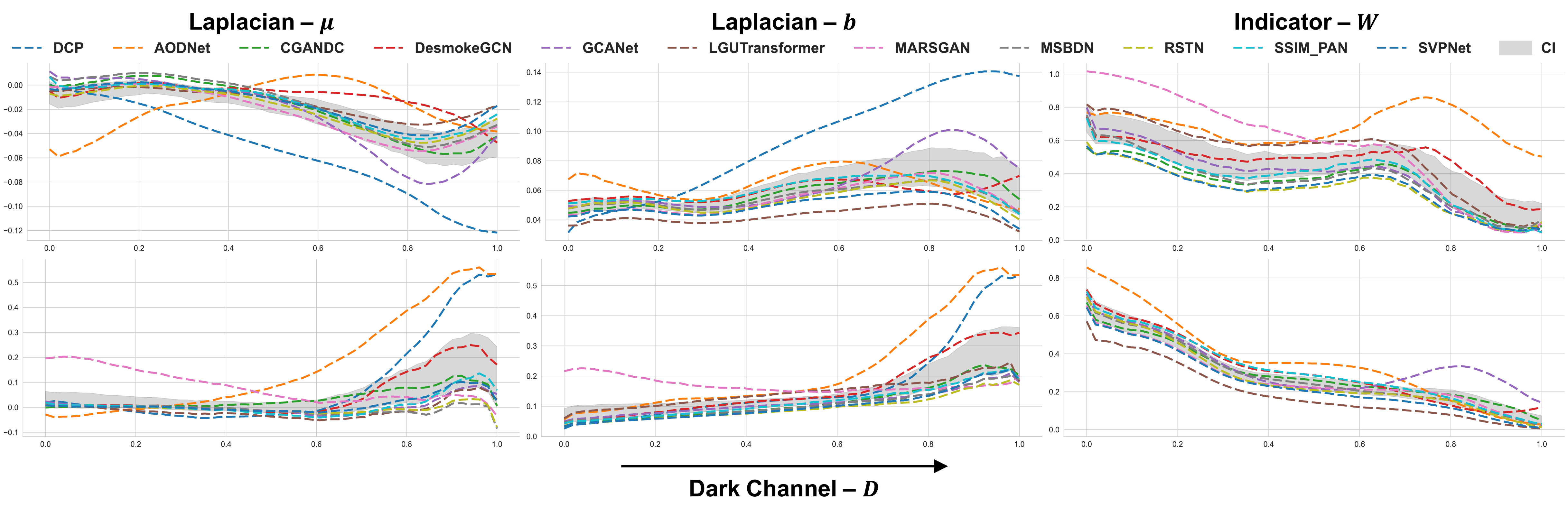}
        \caption{The Laplacian parameters (i.e., $\mu$ and $b$) estimated from the specific dataset, and indicator $W^*$ calculated with Eq.~\eqref{Equ_Method_EMM_MoE_ERR_OPT_SOLUTION}, are presented across different methods with the Confidence Interval (CI). The upper and lower rows correspond to Cholec80 and VASTT-desmoke, respectively.}
        \label{Fig_Laplacian_mu_b_w}
        \end{figure*}
        \textcolor{black}{Interestingly, across the evaluated methods, most produce error distributions that are closer to a Laplacian distribution than to a Gaussian distribution, as shown in \Cref{Fig_Natural_MIS_Error_Comparison} and reported in \Cref{Table_VASST_Cholec_JS_Gau_Lap_Comp}, where most methods yield lower Jensen-Shannon (JS) divergence to the Laplacian model.
        Although this tendency is not uniformly observed across every method and dataset, it represents an empirical statistical trend, particularly on the real-world VASST-desmoke benchmark. 
        We hypothesize this behavior is related to the non-uniform and temporally incoherent nature of real-world surgical smoke, which produces extensive low-error regions in smoke-free areas together with relatively large restoration errors around smoke-contaminated pixels.
        Motivated by this empirical observation of Laplacian-like tendency, we adopt a Laplacian distribution as an approximate statistical prior, that is: }
        \begin{flalign}
            \label{Equ_Method_EMM_DNN_DCP_ERR_DIS_1}
            &\ \varepsilon_\mathcal{M}(x, c) \overset{\mathrm{i.i.d.}}{\sim} \mathrm{Laplace} \left( \mu_\mathcal{M}(x), b_\mathcal{M}(x) \right), &
        \end{flalign}
        where $\mu$ and $b$ are the location and scale parameters, respectively, and $\mathcal{M}$ represents a method, either DCP or DNN.
        Collecting Eqs.\eqref{Equ_Method_EMM_MoE_ERR}, and \eqref{Equ_Method_EMM_DNN_DCP_ERR_DIS_1}, we have:
        \begin{flalign}
            \label{Equ_Method_EMM_MoE_SIGMA}
            &\ \begin{aligned}
            \mathrm{Var}[\varepsilon_\mathrm{MoE}] &= 2 \cdot \big( W^2(x) \cdot b^2_\mathrm{DCP}(x) \\
             & \quad + (1 - W(x))^2 \cdot b^2_\mathrm{DNN}(x) \big),
            \end{aligned} &
        \end{flalign}
        \begin{flalign}
             \label{Equ_Method_EMM_MoE_MU}
            &\ \begin{aligned}
            \mathbb{E}[\varepsilon_\mathrm{MoE}] &= W(x) \cdot \mu_\mathrm{DCP}(x) \\
             & \quad + (1 - W(x)) \cdot \mu_\mathrm{DNN}(x).
            \end{aligned} &
        \end{flalign}
        Then, the original optimization problem Eq.~\eqref{Equ_Method_EMM_MoE_ERR_OPT} is rewritten as:
        \begin{flalign}
            \label{Equ_Method_EMM_MoE_ERR_OPT_2}
            &\ \min_{W(x)} \; \; \mathrm{Var}(\varepsilon_\mathrm{MoE}) + \left( \mathbb{E}(\varepsilon_\mathrm{MoE}) \right)^2, & \\
            \label{Equ_Method_EMM_MoE_ERR_OPT_ST1}
            &\ s.t. \; \; 0 \leq W(x) \leq 1. &
        \end{flalign}
        We solve Eq.~\eqref{Equ_Method_EMM_MoE_ERR_OPT_2} and obtain the unconstrained solution:
        \begin{flalign}
            \label{Equ_Method_EMM_MoE_ERR_OPT_SOLUTION}
            &\ W^*(x) = 
            \frac{2 b^2_{\mathrm{2}, x} - \mu_{\mathrm{2}, x} \cdot (\mu_{\mathrm{1}, x} - \mu_{\mathrm{2}, x})}
            {2b^2_{\mathrm{1}, x} + 2b^2_{\mathrm{2}, x} + (\mu_{\mathrm{1}, x} - \mu_{\mathrm{2}, x})^2}, &
        \end{flalign}
        where $W^*(x)$ is the optimized $W(x)$ without constraints. For simplicity, $b_\mathrm{DCP}(x)$, $b_\mathrm{DNN}(x)$, $\mu_\mathrm{DCP}(x)$ and $\mu_\mathrm{DNN}(x)$ are written as $b_{1, x}$, $b_{2, x}$, $\mu_{1, x}$ and $\mu_{2, x}$, respectively. Subsequently, $W^*(x)$ is clipped to the interval $[0, 1]$ to satisfy the constraint Eq.~\eqref{Equ_Method_EMM_MoE_ERR_OPT_ST1}. \textcolor{black}{Given that several methods exhibit Gaussian-like error distribution (see \Cref{Table_VASST_Cholec_JS_Gau_Lap_Comp}), we further derive the optimized $W(x)$ under Gaussian assumption in \ref{app_gaussian_w} for comparison.}

     
    \subsection{A New Restoration Formula}
        \textcolor{black}{However, Eq.~\eqref{Equ_Method_EMM_MoE_ERR_OPT_SOLUTION} cannot be directly applied in practice due to the unknown Laplacian parameters. 
        To address this issue, we investigate whether the optimal gating weight $W^*$ can be approximated using observable image characteristics. 
        Intuitively, the gating weight should be related to the smoke density or the transmission map, because under dense smoke conditions, traditional DCP may become less effective due to severe information degradation, whereas deep learning methods with stronger reconstruction capabilities may provide a relatively more effective alternative.
        However, neither smoke density nor the transmission map is directly accessible, and thus the dark channel remains an observable proxy \citep{Kaiming_2011_Single_Image}.}  


        \textcolor{black}{To this end, we investigate the correlation between dark channel and gating weights by following Eq.~\eqref{Equ_Method_EMM_MoE_ERR_OPT_SOLUTION}. \Cref{Fig_Laplacian_mu_b_w} presents the trends of $W^*$ given $D$, based on statistics across all methods in two datasets: Cholec80 \citep{Twinanda_2017_EndoNet_A} with synthetic smoke, and the benchmark VASST-desmoke \citep{Xia_2024_A_New, Xia_2025_In_Vivo} containing real-world smoke.}
        \textcolor{black}{Deep learning models display varied trends, while an overall tendency for $W^*$ to decrease as $D$ increases can be observed across methods and datasets. 
        The observed correlationship between $W^*$ and $D$ motivates the following heuristic design:}
        \begin{flalign}
            \label{Equ_Method_RF_WD}
            &\ \hat{W}(x) = 1 - D(x), &
        \end{flalign}
        \begin{table}[t]
            \caption{Pearson Correlation Coefficient between $W$ and $D$.}
            \label{Table_VASST_Cholec_W_D_Pearson}
            \centering
            \resizebox{0.5\textwidth}{!}{
            \begin{tabular}[width=1.0\linewidth]{l c c }
                \toprule
                \textbf{Method} & \textbf{VASST-desmoke} & \textbf{Cholec80} \\
                
                \midrule
                
                AODNet \citep{Li_2017_AOD_Net} & -0.065 & -0.962 \\
                
                CGAN-DC \citep{Salazar_Colores_2020_Desmoking_Laparoscopy} & -0.820 & -0.963 \\
                
                De-smokeGCN \citep{Chen_2020_De_smokeGCN} & -0.763 & -0.973 \\
                
                GCANet \citep{Chen_2019_Gated_Context} & -0.933 & -0.765 \\
                
                LGUTransformer \citep{Wang_2024_Desmoking_of} & -0.899 & -0.954 \\
                
                MARS-GAN \citep{Hong_2023_MARS_GAN} & -0.977 & -0.942 \\
                
                MSBDN \citep{Dong_2020_Multi_Scale} & -0.899 & -0.964 \\
                
                RSTN \citep{Wang_2023_Surgical_smoke} & -0.896 & -0.945 \\
                
                SSIM-PAN \citep{Sidorov_2020_Generative_Smoke} & -0.842 & -0.982 \\
                
                SVPNet \citep{Wang_2024_Smoke_veil} & -0.873 & -0.965 \\
                
                \midrule

                Avg. & -0.797 & -0.942\\
                
                \bottomrule
            \end{tabular}
            }
        \end{table}
        where $\hat{W}$ represents the empirical estimation of $W$ given $D$. The motivations are as follows: 
        \begin{enumerate}
            \item We calculate the Pearson Correlation Coefficient (PCC) between $W$ and $D$ for all methods. Most PCCs are close to $-0.9$, indicating a negative linear correlation, reported in \Cref{Table_VASST_Cholec_W_D_Pearson}. For the low PCC of AODNet in VASST-desmoke, we observe that the original AODNet struggles to deal with dense smoke (i.e., $D$ $\rightarrow$ 1), showing similar quantitative performance to DCP (i.e., $W$ $\rightarrow$ 0.5 means equal significance for DCP and AODNet); this is because AODNet is featured with only five convolutional layers and exhibits weaker reconstruction ability facing dense smoke and fewer training data compared to Cholec80.
            \item Developing a general \textcolor{black}{and compatible} mechanism for the existing DNN methods is one of the objectives of this study, and this function captures the overall correlation between $W$ and $D$; 
            \item Equation~\eqref{Equ_Method_RF_WD} is intuitive: a high intensity of dark channel (i.e., $D(x) \rightarrow 1$) statistically indicates dense smoke \citep{Kaiming_2011_Single_Image, Xia_2025_In_Vivo}, which challenges DCP due to information loss and division by zero or very small values, thereby making DNN the dominant model (i.e., $W(x) \rightarrow 0$); 
            \item It encourages consistent predictions in smokeless regions (i.e., $D(x) \rightarrow 0$); 
            \item The denominator $1 - D$ and the unknown parameter $A$ in Eq.~\eqref{Equ_Method_BG_DCP_RES} are eliminated in the subsequent derivation, stabilizing the computation, simplifying the modeling, and precluding hard value clipping.
        \end{enumerate}
 
        


        By combining Eqs.\eqref{Equ_Method_BG_DCP_RES}, \eqref{Equ_Method_EMM_MoE_1}, and \eqref{Equ_Method_RF_WD}, we obtain: 
        \begin{flalign}
        &\ \begin{aligned}
            \label{Equ_Method_RF_RES_1}
            J_\mathrm{MoE}(x, c) &= 
            I(x, c) - A(c) \cdot D(x) \\
            & \quad + D(x) \cdot J_\mathrm{DNN}(x, c).
        \end{aligned} &
        \end{flalign}
        Then, we use the physical reflection model Eq.~\eqref{Equ_Method_BG_ATM_J} to further decompose $J_\mathrm{DNN}$: 
        \begin{flalign}
        &\ \begin{aligned}
            \label{Equ_Method_RF_RES_2}
            J_\mathrm{MoE}(x, c) &= 
            I(x, c) - A(c) \cdot D(x) \\
            & \quad + D(x) \cdot A(c) \cdot \rho_\mathrm{DNN}(x, c),
        \end{aligned} &
        \end{flalign}
        where $\rho_\mathrm{DNN}(x, c)=J_\mathrm{DNN}(x, c) / A(c)$ represents the normalized radiance obtained from Eq.~\eqref{Equ_Method_BG_ATM_J}. Equation~\eqref{Equ_Method_RF_RES_2} can be reformulated as:
        \begin{flalign}
        &\ \begin{aligned}
            \label{Equ_Method_RF_RES_3}
            J_\mathrm{MoE}(x, c) &= 
            I(x, c) - D(x) \cdot A(c) \cdot (1 - \rho_\mathrm{DNN}(x, c)) \\
            & = I(x, c) - \mathcal{D}(x) \cdot  (1 - \rho_\mathrm{DNN}(x, c)),
        \end{aligned} &
        \end{flalign}
        where $\mathcal{D}$ denotes the denormalized dark channel $D$. Recalling Eq.~\eqref{Equ_Method_BG_DCP}, $\mathcal{D}$ is defined as: 
        \begin{flalign}
            \label{Equ_Method_RF}
            &\ \mathcal{D}(x) 
            = D(x) \cdot A(c) 
            = \underset{c \in C}{\mathrm{min}} \left( \underset{u \in \Omega(x; z)}{\mathrm{min}} \left( I(u, c) \right) \right), &
        \end{flalign}
        where $A$ is eliminated, thereby avoiding the challenging estimation of global light illuminance in laparoscopic surgical scenes.
        In this case, the original DNN modeling in Eqs.\eqref{Equ_Method_BG_DNN_RES} and \eqref{Equ_Method_BG_DNN_GENERAL_LOSS} can be improved as:
        \begin{flalign}
            \label{Equ_Method_RF_DNN_RES}
            &\ \rho_\mathrm{DNN}(x, c) = \sigma \left( \Phi_\theta \left( I(x, c), \tilde{Z} \right) \right), & \\
            \label{Equ_Method_RF_DNN_GENERAL_LOSS}
            &\ \theta^* = \underset{\theta}{\arg \min} \; \; \mathcal{L} \left( J, (I - \mathcal{D} \cdot (1 - \rho_\mathrm{DNN})); \mathcal{X}, \theta \right), &
        \end{flalign}
        where $\sigma(\cdot)$ is a differentiable function (typically a sigmoid) that maps the DNN output to the interval $[0, 1]$. It can be ignored if the output is already constrained within this range, depending on the specific model design. The proposed formulation introduces only minimal modifications to the original DNN model $\Phi_\theta$, without adding any additional trainable parameters. Notably, SurgiATM leverages the denormalized dark channel as a key element, in line with recent methods \citep{Chen_2019_Gated_Context, Hong_2023_MARS_GAN, Chen_2020_De_smokeGCN, Pan_2022_DeSmoke_LAP, Salazar_Colores_2020_Desmoking_Laparoscopy}, suggesting that the dark channel prior plays a useful role in surgical desmoking \citep{Xia_2025_In_Vivo}. \textcolor{black}{Unlike previous methods such as using the dark channel as an auxiliary input feature or a supervision signal for the generator in GAN-based methods \citep{Pan_2022_DeSmoke_LAP, Salazar_Colores_2020_Desmoking_Laparoscopy}, SurgiATM directly incorporates the dark channel into an explicit output-layer formulation, enabling lightweight and compatible integration with different backbone networks.}
    
    \begin{figure*}[t]
    \centering
    \tcbox[colframe=red, colback=white, boxrule=0.25mm, size=tight]{\includegraphics[width=1.0\linewidth]{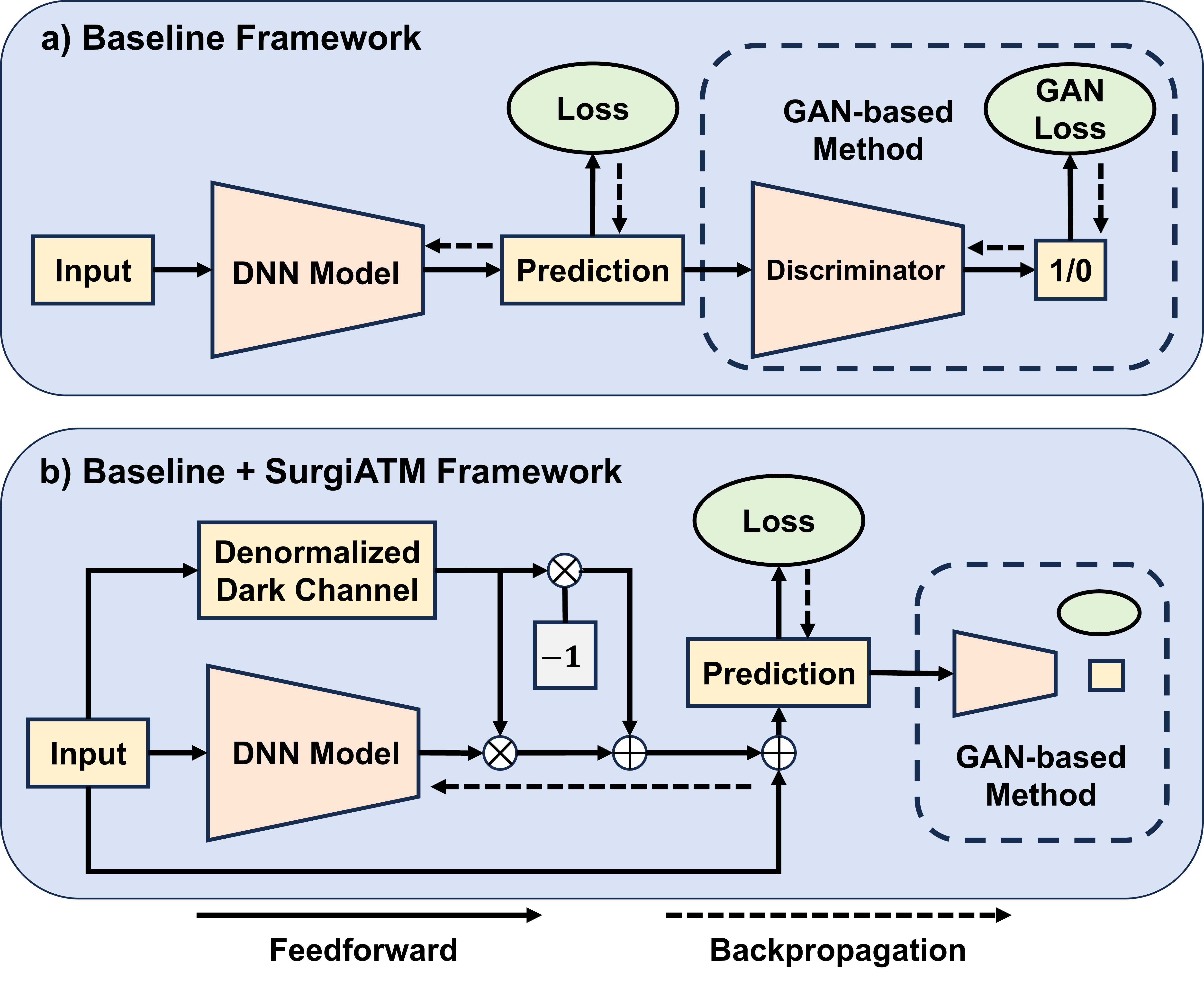}}
    \caption{\textcolor{black}{(a) and (b) display general baseline frameworks for GAN and non-GAN-based methods. (c) and (d) show how to integrate SurgiATM into them. (e) shows some inappropriate usages of SurgiATM; such integration cannot work properly; for simplicity, backpropagation is not marked in (e). (f) demonstrates the internal workflow of SurgiATM.}}
    \label{Fig_Plug_in_baseline}
    \end{figure*}

    \subsection{Gradient Computation and Refinement}
    \label{Sec_grad_comp_ref}
        In this section, we compute the gradient with respect to the model prediction $\rho_\mathrm{DNN}$ for two commonly used loss functions: Mean Absolute Error (MAE or L1) loss and Mean Squared Error (MSE or L2) loss. Their gradients are denoted as $\nabla_\rho \mathcal{L}_1$ and $\nabla_\rho \mathcal{L}_2$, respectively: 
        \begin{flalign}
            \label{Equ_Method_GR_L1_GRAD}
            &\ \nabla_\rho \mathcal{L}_1 = 
            \begin{cases}
                - \mathcal{D}, & J > I - \mathcal{D} \cdot (1 - \rho) ; \\
                \mathcal{D}, & J < I - \mathcal{D} \cdot (1 - \rho) ; \\
                0, & J = I - \mathcal{D} \cdot (1 - \rho),
            \end{cases} &
        \end{flalign}
        \begin{flalign}
            \label{Equ_Method_GR_L2_GRAD}
            &\ \nabla_\rho \mathcal{L}_2 = - \mathcal{D} \cdot (J - (I - \mathcal{D} \cdot (1 - \rho))), &
        \end{flalign}
        where, for simplicity, $\rho_\mathrm{DNN}$ is written as $\rho$. A potential issue arises when $\mathcal{D}(x) = 0$, as this results in a zero gradient, preventing the DNN model from updating its weights at those pixels. To address this, we refine $\mathcal{D}$ as:
        \begin{flalign}
            \label{Equ_Method_GR_D_Refine}
            &\ \hat{\mathcal{D}} = \frac{\mathcal{D} + \eta}{1 + \eta}, \quad \eta \in [0, +\infty), &
        \end{flalign}
        where $\eta$ is a smoothing factor; $\eta = 0$ corresponds to no refinement; and $\eta \rightarrow +\infty$ implies $\hat{\mathcal{D}} \rightarrow 1$, causing Eq.~\eqref{Equ_Method_RF_DNN_GENERAL_LOSS} to degenerate into conventional residual prediction. Ablation results for $\eta$ and $z$ are reported in Section~\ref{Sec_Dis_Ablation_Different_Hyper}.

    \subsection{Plug into Baselines}
        SurgiATM can be regarded as an \textcolor{black}{output layer} appended to the end of DNN models (see \Cref{Fig_Plug_in_baseline}). It introduces only minimal modification to the original baseline. In conventional settings, the output of a DNN model is directly treated as desmoked results, and then losses are computed, and gradients are propagated during training (see \Cref{Fig_Plug_in_baseline}(a) and (b)). 
        
    
        With SurgiATM, the DNN output is first interpreted as a normalized radiance $\rho_\text{DNN}$, and then passed into SurgiATM. Losses are computed using the reconstructed results produced by SurgiATM, and gradients are backpropagated through SurgiATM to the baseline DNN models (see \Cref{Fig_Plug_in_baseline}(c) and (d)). 
        For GAN-based approaches, after incorporating SurgiATM, the discriminator should no longer operate directly on the final output of the generator; instead, it should take the SurgiATM-processed results as input (see \Cref{Fig_Plug_in_baseline}(d)). 
        \textcolor{black}{SurgiATM cannot be used to process intermediate feature maps within the baseline backbone (see \Cref{Fig_Plug_in_baseline}(e)), as it is specifically designed as an output layer at the end of the backbone models, receiving smoky observation and normalized radiance prediction as inputs.}
        \textcolor{black}{Notably, SurgiATM alters the semantic information predicted by the baseline methods, and thus the baselines should be retrained after using SurgiATM.}
        
        On the other hand, Diffusion-based methods have recently emerged as a prevalent paradigm in generative AI, \textcolor{black}{and have also been explored for medical image analysis, including the surgical desmoking task \citep{Amirhossein_2023_Diffusion_Models}.} However, SurgiATM does not currently support diffusion-based approaches. This is because SurgiATM is developed for methods that directly predict smokeless images in a single forward pass and is derived from classic physical models that formulate a direct mapping from smoky frames to smokeless ones. This formulation is fundamentally different from Diffusion models, in which DNNs are charged with predicting Gaussian noise components to be removed (i.e., learning a mapping from noisy images to Gaussian noise) through an iterative denoising process involving multiple inference steps. 
        \textcolor{black}{We analyze and discuss this limitation in Section~\ref{Integration_Conflict}.}

\begin{table*}[t]
    \caption{Training and evaluation on VASST-desmoke (real smoke). There is no overlap between training and evaluation sets. The results are represented in the form of "Baseline / Baseline + SurgiATM". \textbf{Bold} represents that baseline is better or that integrated with SurgiATM is better. \textcolor{black}{"$\downarrow$" indicates that lower values are better. "$\uparrow$" means higher is better.}}
    \label{Tab_VASST_Intra_Dom}
    \centering
    \resizebox{1.0\textwidth}{!}{
    \begin{tabular}{l c c c c c c c c}
        \toprule
        Method & Backbone & Loss Function & CIEDE2000 $\downarrow$ & PSNR $\uparrow$ & RMSE $\downarrow$ & SSIM $\uparrow$ \\ 
        \midrule 
        AODNet \citep{Li_2017_AOD_Net} & CNN & MSE & 10.349 / \textbf{6.327} & 17.378 / \textbf{21.994} & 0.147 / \textbf{0.090} & 0.693 / \textbf{0.789} \\ 
        GCANet \citep{Chen_2019_Gated_Context} & CNN & MSE & 7.893 / \textbf{5.369} & 21.091 / \textbf{23.907} & 0.095 / \textbf{0.073} & 0.729 / \textbf{0.806} \\ 
        De-smokeGCN \citep{Chen_2020_De_smokeGCN} & U-Net & MAE + Others & 9.577 / \textbf{7.348} & 19.779 / \textbf{22.127} & 0.115 / \textbf{0.094} & 0.550 / \textbf{0.717} \\ 
        MSBDN \citep{Dong_2020_Multi_Scale} & U-Net & MSE & 6.679 / \textbf{5.164} & 21.604 / \textbf{23.881} & 0.088 / \textbf{0.073} & 0.651 / \textbf{0.813} \\ 
        SVPNet \citep{Wang_2024_Smoke_veil} & U-Net & MSE + Others & 5.574 / \textbf{5.141} & 23.464 / \textbf{24.362} & 0.073 / \textbf{0.070} & 0.789 / \textbf{0.820} \\ 
        CGAN-DC \citep{Chen_2019_Gated_Context} & cGAN & Adversarial + Others & 6.735 / \textbf{5.812} & 22.453 / \textbf{23.809} & 0.083 / \textbf{0.079} & 0.676 / \textbf{0.782} \\ 
        MARS-GAN \citep{Hong_2023_MARS_GAN} & CycleGAN & Adversarial + Others & 6.389 / \textbf{5.899} & 20.121 / \textbf{23.488} & 0.113 / \textbf{0.076} & 0.602 / \textbf{0.819} \\ 
        SSIM-PAN \citep{Sidorov_2020_Generative_Smoke} & GAN & Adversarial + Others & 7.046 / \textbf{5.811} & 21.465 / \textbf{23.477} & 0.090 / \textbf{0.075} & 0.655 / \textbf{0.750} \\ 
        RSTN \citep{Wang_2023_Surgical_smoke} & SwinT & Charbonnier + Others & 5.385 / \textbf{4.976} & 23.263 / \textbf{24.306} & 0.074 / \textbf{0.071} & 0.750 / \textbf{0.824} \\ 
        LGUTransformer \citep{Wang_2024_Desmoking_of} & SwinT + U-Net & Charbonnier & 9.927 / \textbf{5.990} & 18.238 / \textbf{23.376} & 0.121 / \textbf{0.080} & 0.510 / \textbf{0.761} \\ 
        \bottomrule 
    \end{tabular}
    }
\end{table*}

\section{Experiments and Results}
    \subsection{Implementation Details}
    \label{Exp_implement_detail}
        For all comparisons, we set the smoothing factor $\eta=0.1$ because this value contributes to better results in our ablation study, and set the window size $z=15$ inherited from DCP \citep{Kaiming_2011_Single_Image}. The frames are resized to 256$\times$256. All experiments are conducted on a 24GB NVIDIA RTX 3090 GPU. SurgiATM is extended as an output layer at the end of all approaches, without any modification to the original architectures, input forms, loss functions, and hyperparameters. To ensure fair comparisons, training configurations are standardized across all methods, employing the Adam optimizer with a learning rate of 0.0002, beta1 of 0.9, beta2 of 0.999, and epochs of 50; \textcolor{black}{both the original model and the SurgiATM-integrated model are independently trained from scratch with the same random initialization (i.e., the same random seed); the SurgiATM-integrated model is not obtained via fine-tuning the original model.} Due to GPU memory constraints, the batch size depends on their respective publications. Please visit \hyperlink{https://github.com/MingyuShengSMY/SurgiATM}{https://github.com/MingyuShengSMY/SurgiATM} for more implementation details. We implement SurgiATM as a \textit{torch.nn.Module} and upload it to the link, which can be directly used. 

    \subsection{Datasets and Metric}
        \subsubsection{Datasets} 
            In this study, we utilize three datasets: Cholec80 \citep{Twinanda_2017_EndoNet_A}, VASST-desmoke \citep{Xia_2024_A_New, Xia_2025_In_Vivo}, and the Hamlyn Centre Laparoscopic and Endoscopic Dataset Videos (referred to as the Hamlyn Dataset for simplicity) \citep{Giannarou_2013_Probabilistic_Tracking, Ye_2017_Self_Supervised}. The Hamlyn Dataset is specifically reserved for external evaluation. 
            For Cholec80, 500 smokeless and 500 smoky frames are randomly sampled from each video \citep{Leibetseder_2017_Real_Time}, with the smoke-free frames then blended with synthetic smoke \citep{Bolkar_2018_Deep_Smoke} for training.
            VASST-desmoke is a recently released paired benchmark for laparoscopic surgical smoke removal. It provides both real smoky frames and corresponding ground-truth smoke-free annotations, thereby enabling evaluation on real surgical smoke with intuitive full-reference metrics. Potentially, VASST-desmoke also allows models to learn from real smoke. In addition, \citet{Xia_2025_In_Vivo} further expand the VASST-desmoke, producing more paired frames sourced from Cholec80. For convenience, we note the latest part of the dataset as "VASST-desmoke-C".
            The Hamlyn Dataset is excluded from all training and statistical analysis procedures in this work, and its frames are sampled from the videos at 25 frames per second (FPS).
            We perform five-fold cross-evaluation, ensuring no video overlap between the training and evaluation sets. 
            The three datasets record various surgical procedures with inconsistent camera devices and lighting environments in different hospitals, significantly challenging the stability and generalizability of surgical desmoking methods.


        \subsubsection{Metrics} 
            For the evaluation of paired datasets, we leverage full-reference metrics: CIEDE2000, PSNR, RMSE, and SSIM, to evaluate perceptual color difference, reconstruction quality, prediction error, and structure similarity, respectively. 
            For the unpaired datasets, we evaluate model performance using non-reference metrics, including BRISQUE and NIQE to measure how close an image is to natural scenes, and FADE \citep{Choi_2015_Referenceless_Prediction}, which specifically measures fog-related degeneration.

    \subsection{Comparison of Quantitative Results}
    \label{Sec_RIDE}
        We train five folds of ten existing methods and their corresponding SurgiATM modified versions, using Cholec80 with synthetic smoke and VASST-desmoke with real-world smoke, respectively. We then compare their five-fold average performance on the real-world surgical smoke datasets: VASST-desmoke, Cholec80 (real smoke), and the Hamlyn Dataset. In total, we train 200 models and conduct 500 evaluations.
        
        
        The quantitative results on VASST-desmoke and VASST-desmoke-C are presented in \Cref{Tab_VASST_Intra_Dom,Tab_CholecVASST,Tab_Cholec_To_VASST}. They demonstrate that incorporating SurgiATM effectively improves model performance, regardless of the backbone architecture or loss function. 
        However, for SVPNet in \Cref{Tab_Cholec_To_VASST}, SurgiATM yields no improvement in SSIM. 
        We argue that, compared to other desmoking methods, SVPNet \citep{Wang_2024_Smoke_veil} employs an energy function \citep{Wang_2018_Variational_based} for restoration. This is a physics-guided approach similar to ours, and its three physics-related parameters are predicted by three independent U-Nets. 
        As a result, incorporating an extra physics-guided component (our SurgiATM) into SVPNet may slightly disrupt the internal physical consistency, leading to the observed deterioration in SSIM.
        
        
        For Cholec80, we report the evaluation results in \Cref{Tab_Cholec_To_Cholec}. Our SurgiATM effectively improves most model performances in terms of the fog-related metric (i.e., FADE score).
        In terms of the external evaluation on the Hamlyn Dataset, as shown in \Cref{Tab_Cholec_To_Hamlyn,Tab_VASST_To_Hamlyn}, for the defogging metric FADE, employing our SurgiATM enhances most methods, except for the traditional dehazing method AODNet in \Cref{Tab_Cholec_To_Hamlyn}.
        We attribute this to two factors: 1) the simple architecture and insufficient trainable weights of AODNet (i.e., only five convolutional layers) compared to other methods, and 2) the non-Laplacian error distribution of AODNet when trained on Cholec80 with synthetic smoke, as reported in \Cref{Table_VASST_Cholec_JS_Gau_Lap_Comp}.
        In contrast, although GCANet and SSIM-PAN also exhibit non-Laplacian error distributions in Cholec80, and CGAN-DC and MARS-GAN in VASST-desmoke, their performances are effectively enhanced owing to both their well-designed architectures and the strong generalizability of SurgiATM.


        \textcolor{black}{We also observe a slight performance gap between training on the VASST-desmoke dataset with real smoke (see \Cref{Tab_VASST_To_Hamlyn}) and training on Cholec80 with synthetic smoke (see \Cref{Tab_Cholec_To_Hamlyn}). Regardless of whether SurgiATM is used, training with real smoke usually yields lower FADE values, reflecting the inherent domain gap between real and synthetic smoke distributions. Although SurgiATM improves the baseline in most settings, it cannot eliminate the performance gap caused by the discrepancy between the two data distributions. These findings empirically suggest that training with real smoke data is more effective.}

        
        On the other hand, SurgiATM demonstrates limited effectiveness in terms of BRISQUE and NIQE across the ten existing methods in \Cref{Tab_Cholec_To_Cholec,Tab_VASST_To_Hamlyn,Tab_Cholec_To_Hamlyn}. The main reason is that these two metrics are designed to measure similarity to natural images, which may not generalize well to surgical endoscopic frames due to the substantial domain gap between them \citep{Wang_2024_Desmoking_of}. The scarcity of domain-relevant non-referenced IQA metrics is exactly an underexplored research problem in the area of surgical smoke removal. However, addressing this limitation is beyond the scope of the current study, and developing reliable, domain-specific metrics for surgical videos remains a promising direction for future research.

    \begin{table}[t]
        \caption{Training and evaluation on VASST-desmoke-C (real smoke).}
            \label{Tab_CholecVASST}     
            \centering
            \resizebox{0.5\textwidth}{!}{
            \begin{tabular}{l c c c c}
                \toprule
                Method & CIEDE2000 $\downarrow$ & PSNR $\uparrow$ & RMSE $\downarrow$ & SSIM $\uparrow$ \\ 
                \midrule 
                AODNet \citep{Li_2017_AOD_Net} & 7.344 / \textbf{5.297} & 20.727 / \textbf{24.655} & 0.098 / \textbf{0.065} & 0.860 / \textbf{0.900} \\ 
                GCANet \citep{Chen_2019_Gated_Context} & 7.453 / \textbf{4.291} & 22.566 / \textbf{26.426} & 0.080 / \textbf{0.053} & 0.812 / \textbf{0.901} \\ 
                De-smokeGCN \citep{Chen_2020_De_smokeGCN} & 6.558 / \textbf{5.118} & 24.340 / \textbf{26.125} & 0.067 / \textbf{0.059} & 0.725 / \textbf{0.837} \\ 
                MSBDN \citep{Dong_2020_Multi_Scale} & 4.861 / \textbf{3.999} & 25.532 / \textbf{27.496} & 0.058 / \textbf{0.051} & 0.834 / \textbf{0.905} \\ 
                SVPNet \citep{Wang_2024_Smoke_veil} & 4.509 / \textbf{3.743} & 26.663 / \textbf{28.121} & 0.052 / \textbf{0.047} & 0.892 / \textbf{0.915} \\ 
                CGAN-DC \citep{Chen_2019_Gated_Context} & 5.746 / \textbf{4.599} & 25.429 / \textbf{26.969} & 0.059 / \textbf{0.054} & 0.782 / \textbf{0.832} \\ 
                MARS-GAN \citep{Hong_2023_MARS_GAN} & 4.863 / \textbf{4.592} & 26.096 / \textbf{27.276} & 0.055 / \textbf{0.051} & 0.888 / \textbf{0.910} \\ 
                SSIM-PAN \citep{Sidorov_2020_Generative_Smoke} & 5.382 / \textbf{4.282} & 25.092 / \textbf{27.003} & 0.061 / \textbf{0.052} & 0.799 / \textbf{0.877} \\ 
                RSTN \citep{Wang_2023_Surgical_smoke} & 4.321 / \textbf{3.532} & 26.361 / \textbf{28.395} & 0.053 / \textbf{0.047} & 0.854 / \textbf{0.915} \\ 
                LGUTransformer \citep{Wang_2024_Desmoking_of} & 16.862 / \textbf{5.538} & 17.866 / \textbf{27.532} & 0.130 / \textbf{0.075} & 0.189 / \textbf{0.895} \\ 
                \bottomrule 
            \end{tabular}
            }
    \end{table}

        \begin{table}[t]
            \caption{Training on Cholec80 (synthetic smoke) and evaluation on VASST-desmoke (real smoke).}
            \label{Tab_Cholec_To_VASST}     
            \centering
            \resizebox{0.5\textwidth}{!}{
            \begin{tabular}{l c c c c}
                \toprule
                Method & CIEDE2000 $\downarrow$ & PSNR $\uparrow$ & RMSE $\downarrow$ & SSIM $\uparrow$ \\ 
                \midrule 
                AODNet \citep{Li_2017_AOD_Net} & 9.609 / \textbf{7.774} & 18.621 / \textbf{21.526} & 0.132 / \textbf{0.109} & 0.729 / \textbf{0.798} \\ 
                GCANet \citep{Chen_2019_Gated_Context} & 9.803 / \textbf{7.042} & 19.338 / \textbf{22.220} & 0.119 / \textbf{0.097} & 0.747 / \textbf{0.801} \\ 
                De-smokeGCN \citep{Chen_2020_De_smokeGCN} & 8.370 / \textbf{7.585} & 20.654 / \textbf{22.088} & 0.109 / \textbf{0.102} & 0.716 / \textbf{0.788} \\ 
                MSBDN \citep{Dong_2020_Multi_Scale} & 8.277 / \textbf{7.327} & 20.360 / \textbf{22.153} & 0.114 / \textbf{0.103} & 0.769 / \textbf{0.810} \\ 
                SVPNet \citep{Wang_2024_Smoke_veil} & 8.105 / \textbf{7.797} & 21.196 / \textbf{21.712} & 0.110 / \textbf{0.109} & \textbf{0.806} / 0.793 \\ 
                CGAN-DC \citep{Chen_2019_Gated_Context} & 8.177 / \textbf{7.525} & 20.663 / \textbf{21.988} & 0.113 / \textbf{0.105} & 0.770 / \textbf{0.805} \\ 
                MARS-GAN \citep{Hong_2023_MARS_GAN} & 7.808 / \textbf{7.444} & 21.444 / \textbf{22.110} & 0.107 / \textbf{0.104} & 0.792 / \textbf{0.811} \\ 
                SSIM-PAN \citep{Sidorov_2020_Generative_Smoke} & 8.350 / \textbf{7.600} & 20.338 / \textbf{21.805} & 0.115 / \textbf{0.106} & 0.761 / \textbf{0.806} \\ 
                RSTN \citep{Wang_2023_Surgical_smoke} & 7.963 / \textbf{7.533} & 21.034 / \textbf{22.065} & 0.111 / \textbf{0.107} & 0.763 / \textbf{0.813} \\ 
                LGUTransformer \citep{Wang_2024_Desmoking_of} & 7.357 / \textbf{7.352} & 22.108 / \textbf{22.111} & 0.105 / \textbf{0.104} & 0.812 / \textbf{0.818} \\ 
                \bottomrule 
            \end{tabular}
            }
        \end{table}
        
        \begin{table}[t]
            \caption{Training using synthetic smoke and evaluation using real smoke on Cholec80 (non-overlap videos).}
            \label{Tab_Cholec_To_Cholec}
            \centering
            \resizebox{0.485\textwidth}{!}{
            \begin{tabular}{l c c c}
                \toprule
                Method & BRISQUE $\downarrow$ & FADE $\downarrow$ & NIQE $\downarrow$ \\ 
                \midrule 
                AODNet \citep{Li_2017_AOD_Net} & \textbf{12.551} / 14.814 & \textbf{0.440} / 0.484 & 10.449 / \textbf{9.434} \\ 
                GCANet \citep{Chen_2019_Gated_Context} & \textbf{7.795} / 11.566 & 0.407 / \textbf{0.403} & 9.866 / \textbf{9.331} \\ 
                De-smokeGCN \citep{Chen_2020_De_smokeGCN} & \textbf{4.857} / 5.575 & 0.384 / \textbf{0.376} & 10.885 / \textbf{9.569} \\ 
                MSBDN \citep{Dong_2020_Multi_Scale} & \textbf{11.838} / 12.176 & 0.421 / \textbf{0.411} & 10.299 / \textbf{9.026} \\ 
                SVPNet \citep{Wang_2024_Smoke_veil} & 12.812 / \textbf{12.134} & 0.416 / \textbf{0.402} & 10.579 / \textbf{9.387} \\ 
                CGAN-DC \citep{Chen_2019_Gated_Context} & 10.258 / \textbf{9.945} & 0.413 / \textbf{0.404} & 9.517 / \textbf{9.039} \\ 
                MARS-GAN \citep{Hong_2023_MARS_GAN} & 13.999 / \textbf{11.680} & 0.416 / \textbf{0.413} & \textbf{13.336} / 13.884 \\ 
                SSIM-PAN \citep{Sidorov_2020_Generative_Smoke} & \textbf{10.539} / 11.093 & 0.411 / \textbf{0.403} & \textbf{9.811} / 11.999 \\ 
                RSTN \citep{Wang_2023_Surgical_smoke} & 13.868 / \textbf{12.652} & 0.420 / \textbf{0.414} & \textbf{12.245} / 16.910 \\ 
                LGUTransformer \citep{Wang_2024_Desmoking_of} & \textbf{9.126} / 11.236 & 0.417 / \textbf{0.416} & 13.734 / \textbf{12.471} \\ 
                \bottomrule 
            \end{tabular}
            }
        \end{table}

\begin{table}[t]
    \caption{Training on VASST-desmoke (real smoke) and evaluation on Hamlyn Dataset (real smoke).}
    \label{Tab_VASST_To_Hamlyn}
    \centering
    \resizebox{0.485\textwidth}{!}{
    \begin{tabular}{l c c c}
        \toprule
        Method & BRISQUE $\downarrow$ & FADE $\downarrow$ & NIQE $\downarrow$ \\ 
        \midrule 
        AODNet \citep{Li_2017_AOD_Net} & 36.549 / \textbf{30.515} & 0.331 / \textbf{0.322} & 12.418 / \textbf{11.851} \\ 
        GCANet \citep{Chen_2019_Gated_Context} & \textbf{13.354} / 26.555 & 0.324 / \textbf{0.294} & 11.512 / \textbf{10.983} \\ 
        De-smokeGCN \citep{Chen_2020_De_smokeGCN} & 33.236 / \textbf{28.007} & 0.329 / \textbf{0.325} & 29.448 / \textbf{18.802} \\ 
        MSBDN \citep{Dong_2020_Multi_Scale} & 26.992 / \textbf{25.198} & 0.384 / \textbf{0.322} & 14.224 / \textbf{10.944} \\ 
        SVPNet \citep{Wang_2024_Smoke_veil} & 39.862 / \textbf{30.325} & 0.401 / \textbf{0.317} & 16.245 / \textbf{13.306} \\ 
        CGAN-DC \citep{Chen_2019_Gated_Context} & \textbf{14.052} / 16.699 & 0.352 / \textbf{0.336} & 16.616 / \textbf{13.216} \\ 
        MARS-GAN \citep{Hong_2023_MARS_GAN} & 33.722 / \textbf{30.601} & 0.377 / \textbf{0.369} & \textbf{15.011} / 16.381 \\ 
        SSIM-PAN \citep{Sidorov_2020_Generative_Smoke} & \textbf{11.936} / 16.047 & 0.357 / \textbf{0.298} & \textbf{16.613} / 20.297 \\ 
        RSTN \citep{Wang_2023_Surgical_smoke} & 35.629 / \textbf{30.232} & 0.412 / \textbf{0.330} & 16.442 / \textbf{12.645} \\ 
        LGUTransformer \citep{Wang_2024_Desmoking_of} & 35.260 / \textbf{29.706} & 0.295 / \textbf{0.294} & 15.300 / \textbf{12.135} \\ 
        \bottomrule 
    \end{tabular}
    }
\end{table}

\begin{table}[t!]
    \caption{Training on Cholec80 (synthetic smoke) and evaluation on Hamlyn Dataset (real smoke).}
    \label{Tab_Cholec_To_Hamlyn}
    \centering
    \resizebox{0.485\textwidth}{!}{
    \begin{tabular}{l c c c}
        \toprule
        Method & BRISQUE $\downarrow$ & FADE $\downarrow$ & NIQE $\downarrow$ \\ 
        \midrule 
        AODNet \citep{Li_2017_AOD_Net} & \textbf{30.292} / 31.168 & \textbf{0.426} / 0.432 & \textbf{11.074} / 11.834 \\ 
        GCANet \citep{Chen_2019_Gated_Context} & \textbf{19.366} / 26.102 & 0.385 / \textbf{0.293} & 11.628 / \textbf{11.083} \\ 
        De-smokeGCN \citep{Chen_2020_De_smokeGCN} & \textbf{13.920} / 17.647 & 0.330 / \textbf{0.315} & 11.628 / \textbf{11.467} \\ 
        MSBDN \citep{Dong_2020_Multi_Scale} & \textbf{26.630} / 27.559 & 0.409 / \textbf{0.332} & 12.152 / \textbf{11.882} \\ 
        SVPNet \citep{Wang_2024_Smoke_veil} & 29.065 / \textbf{28.375} & 0.406 / \textbf{0.267} & 14.358 / \textbf{12.292} \\ 
        CGAN-DC \citep{Chen_2019_Gated_Context} & 25.175 / \textbf{24.127} & 0.408 / \textbf{0.315} & \textbf{10.445} / 10.741 \\ 
        MARS-GAN \citep{Hong_2023_MARS_GAN} & 29.811 / \textbf{27.287} & 0.409 / \textbf{0.311} & 16.443 / \textbf{15.136} \\ 
        SSIM-PAN \citep{Sidorov_2020_Generative_Smoke} & \textbf{24.605} / 26.001 & 0.395 / \textbf{0.312} & \textbf{10.852} / 19.337 \\ 
        RSTN \citep{Wang_2023_Surgical_smoke} & \textbf{26.305} / 28.347 & 0.414 / \textbf{0.328} & \textbf{13.875} / 14.217 \\ 
        LGUTransformer \citep{Wang_2024_Desmoking_of} & \textbf{23.974} / 26.497 & 0.383 / \textbf{0.346} & \textbf{16.786} / 17.448 \\ 
        \bottomrule 
    \end{tabular}
    }
\end{table}

    \begin{figure*}[t]
    \centering
    \includegraphics[width=1\linewidth]{Ablation.pdf}
    \caption{Each heatmap illustrates the ablation study for a specific baseline model incorporated with SurgiATM, showing the impact of different hyperparameter settings in terms of the RMSE metric (lower is better). Each heatmap uses an independent color scale.}
    \label{Fig_Ablation}
    \end{figure*}

    \subsection{Comparison of Qualitative Results}
    
        In this section, we present qualitative comparisons across several representative baselines: De-smokeGCN, GCANet, MARS-GAN, and SVPNet, each selected for different reasons. Specifically, De-smokeGCN \citep{Chen_2020_De_smokeGCN} is a classic DNN-based desmoking method introduced in 2020; GCANet \citep{Chen_2019_Gated_Context} is designed for natural dehazing and deraining; MARS-GAN represents an adversarial-learning-based approach; and SVPNet \citep{Wang_2024_Smoke_veil}, published in 2024, uses a physics-based loss function tailored for surgical desmoking. We train the models on Cholec80 with synthetic smoke and evaluate them on VASST-desmoke, Cholec80, and the Hamlyn Dataset with real surgical smoke, challenging their stability and generalizability.
    
    
        As shown in the error maps in \Cref{Fig_Cholec_to_VASST}, SurgiATM effectively reduces desmoking errors in a computationally efficient and interpretable manner, making it well-suited for deployment on edge devices and for stable surgical desmoking. For the natural dehazing method GCANet, it causes color distortion in \Cref{Fig_Cholec_to_VASST}(a), which is corrected to a certain degree by using SurgiATM. In addition, the proposed approach can mitigate the grid artifact (see the first two rows in \Cref{Fig_Cholec_to_VASST}), demonstrating promising stability. 
        As shown in \Cref{Fig_Cholec80_smoky_and_to_Hamlyn}, the desmoking results of the baselines are improved to varying degrees. For instance, De-smokeGCN and GCANet can produce more stable results with SurgiATM by reducing the grid artifacts (see \Cref{Fig_Cholec80_smoky_and_to_Hamlyn}(a)-(d) of De-smokeGCN) and correcting the restoration colors (see \Cref{Fig_Cholec80_smoky_and_to_Hamlyn}(d) of GCANet). Moreover, SurgiATM helps baselines eliminate more smoke, as shown in \Cref{Fig_Cholec80_smoky_and_to_Hamlyn}(a) for MARS-GAN and \Cref{Fig_Cholec80_smoky_and_to_Hamlyn}(c) for SVPNet, GCANet, and MARS-GAN.


        \textcolor{black}{With the failure cases highlighted by the red boxes in \Cref{Fig_Cholec_to_VASST,Fig_Cholec80_smoky_and_to_Hamlyn}, we observe that SurgiATM may occasionally introduce degraded regions around dark pixels. This behavior arises because the proposed weighting mechanism assigns lower weights to pixels with low dark-channel intensity, thereby reducing the gradient magnitude in these regions, as analyzed in Section~\ref{Sec_grad_comp_ref}. Although the smoothing factor $\eta$ is introduced to alleviate this problem, it cannot completely resolve it.
        This reflects an inherent limitation of SurgiATM. The proposed formulation is designed based on the observation that surgical smoke is generally more prominent in brighter regions, while darker regions are more likely to correspond to smoke-free tissues or shadows. Consequently, when dense smoke appears in dark regions or when anatomical structures exhibit very low intensity, the dark-channel-guided weighting may underestimate their contributions, resulting in local restoration degradation.
        Addressing this challenge may require a more accurate characterization of brightness variations and their underlying physical properties in surgical scenes, which constitutes an interesting direction for future research.}

\section{Discussion}
\label{discuss}

    \subsection{Ablation Study}
    \label{Sec_RAS}

        We conduct ablation studies to investigate the influence of various factors, including smoothing factor $\eta$ and shifting window size $z$, on model performance. Experiments are carried out on the real-world benchmark VASST-desmoke using three representative baselines (i.e., AODNet, RSTN, and SSIM-PAN), chosen for their minimal reliance on auxiliary modules and loss functions, thereby enabling a clear assessment of the impact. All ablation results are averaged from five-fold cross-validation. All experiments follow the same configurations introduced in Section~\ref{Exp_implement_detail}. The only differences are specifically demonstrated in the following section.

        \subsubsection{Different Hyperparameters}
        \label{Sec_Dis_Ablation_Different_Hyper}
    
        
        As shown in \Cref{Fig_Ablation}, we observed that, for most methods, $\eta=0.1$ usually leads to better performance, a trend that is obviously observed for AODNet. This hyperparameter defines a lower bound of the denormalized dark channel $\mathcal{D}$ (see Eq.~\eqref{Equ_Method_GR_D_Refine}) and consequently sets the minimum gradient propagated from loss functions (see Eqs.~\eqref{Equ_Method_GR_L1_GRAD} and \eqref{Equ_Method_GR_L2_GRAD}). \textcolor{black}{Zero and overly small values for smoother (e.g., 0.05, 0.01, or 0.001) may contribute to gradient vanishing, which hinders model fitting and updating and usually demonstrates moderately inferior performance}. Conversely, large values allow higher-magnitude gradients, but the effectiveness of $\mathcal{D}$ is diminished (see Eq.~\eqref{Equ_Method_GR_D_Refine}), gradually degenerating into general residual learning.
        
        
        Another potential trend is observed with respect to the window size $z$: larger window sizes are preferred for AODNet and SSIM-PAN, whereas smaller ones perform better for RSTN. Analyzing their architectures, we infer that the CNN backbones used by AODNet and SSIM-PAN, which have comparatively smaller receptive fields than the SwinT backbone used by RSTN, benefit from a larger window size to capture visual information over a wider range. 
        

        Therefore, when using SurgiATM, $\eta=0.1$ is recommended as a default configuration, while the optimal window size should be selected based on the specific baseline method. 


        \subsubsection{\textcolor{black}{Different Gating Strategies}}
        \label{Sec_Dis_Ablation_Different_Gate}

        \begin{table}[t]
            \caption{\textcolor{black}{ Experimental results of different gating strategies. \textbf{Bold} indicates the best metric values for each baseline. \underline{Underline} marks the second-best ones. "*" denotes the third-best ones. "Null" denotes the original baseline. "Ours" represents the baseline augmented with SurgiATM. "LL" indicates that the implemented gating weights are defined by Eq.~\eqref{Equ_Method_EMM_MoE_ERR_OPT_SOLUTION}, and the parameters of the Laplacian distributions are estimated using a three-layer CNN. "LW" denotes that the gating weights are directly predicted by a three-layer CNN instead of being computed from any prior formulations. "W5" indicates that the gating weight is fixed at 0.5 for all pixels. Since some values are very close to one another, this table reports results to four decimal places.}}
            \label{Tab_ablation_W}
            \centering
            \resizebox{0.5\textwidth}{!}{
            \tcbox[colframe=red, colback=white, boxrule=0.25mm, size=tight]{\begin{tabular}{l c c c c c}
                \toprule
                Method & Gating & CIEDE2000 $\downarrow$ & PSNR $\uparrow$ & RMSE $\downarrow$ & SSIM $\uparrow$ \\ 
                \midrule
                \multirow{5}{*}{AODNet \citep{Li_2017_AOD_Net}} & Null & 10.3495 & 17.3776 & 0.1475 & 0.6930 \\ 
                ~ & Ours & \textbf{6.3270} & \textbf{21.9940} & \textbf{0.0900} & \textbf{0.7890} \\ 
                ~ & LL & 7.5765* & 21.0023* & 0.0963* & 0.7613* \\ 
                ~ & LW & \underline{7.4899} & \underline{21.1517} & \underline{0.0961} & \underline{0.7671} \\ 
                ~ & W5 & 7.5967 & 20.9967 & 0.0973 & 0.7611 \\ 
                \midrule
                \multirow{5}{*}{RSTN \citep{Wang_2023_Surgical_smoke}} & Null & 5.3848 & 23.2633 & 0.0736 & 0.7504 \\ 
                ~ & Ours & \textbf{4.9760} & \textbf{24.3060} & \textbf{0.0710} & \textbf{0.8240} \\ 
                ~ & LL & 5.3171* & \underline{23.9038} & \underline{0.0712} & \underline{0.8127} \\ 
                ~ & LW & \underline{5.3140} & 23.8508* & 0.0713* & 0.8099* \\ 
                ~ & W5 & 5.3551 & 23.7564 & 0.0726 & 0.8091 \\ 
                \midrule
                \multirow{5}{*}{SSIM-PAN \citep{Sidorov_2020_Generative_Smoke}} & Null & 7.0460 & 21.4649 & 0.0898 & 0.6555 \\ 
                ~ & Ours & \textbf{5.8110} & \textbf{23.4770} & \textbf{0.0750} & \textbf{0.7500} \\ 
                ~ & LL & \underline{5.9762} & \underline{22.9245} & \underline{0.0776} & 0.7455* \\ 
                ~ & LW & 6.1036* & 22.8544* & 0.0795* & \underline{0.7480} \\ 
                ~ & W5 & 6.2874 & 22.6948 & 0.0813 & 0.7394 \\ 
                \bottomrule
            \end{tabular}
            }
            }
        \end{table}
        \textcolor{black}{
        Regarding Eq.~\eqref{Equ_Method_RF_WD}, we conduct extensive ablation studies using various gating formulations. The results are summarized in \Cref{Tab_ablation_W}.
        }


        \textcolor{black}{
        Comparing "Null" and "W5" in \Cref{Tab_ablation_W}, we observe that although the gating weight is fixed at 0.5, incorporating the dark channel prior (DCP), i.e., using Eq.~\eqref{Equ_Method_EMM_MoE_1}, produces better results.
        Furthermore, "LW" consistently outperforms "W5", indicating that the mixture-of-experts (MoE) gating mechanism plays an effective role.
        "LL" and "LW" exhibit very similar performance, making it difficult to determine which is superior. This is expected because they share similar optimization objectives. For "LL" (see Eq.~\eqref{Equ_Method_EMM_MoE_ERR_OPT_SOLUTION}), the parameters of the Laplacian distributions are predicted by a CNN and are subsequently used to compute the gating weights, resulting in an optimization objective similar to that of "LW", which directly predicts the gating weights.
        }


        \textcolor{black}{
        Observing the "Ours" and "LW" rows, the trainable-parameter-free SurgiATM based on the prior formulation in Eq.~\eqref{Equ_Method_RF_WD} outperforms the learnable gating strategy, confirming the effectiveness of the physics-guided formulation. 
        In this experiment, a lightweight CNN was designed to predict the gating weights, with the architecture: "Conv(4, 128) $\rightarrow$ ReLU $\rightarrow$ Conv(128, 128) $\rightarrow$ ReLU $\rightarrow$ Conv(128, 1)", where all convolutional layers use a kernel size of 3, a stride of 1, padding of 1, and its inputs are RGB and dark channel.
        Although "LW" performs worse than SurgiATM, we do not exclude the possibility that more sophisticated learnable gating architectures, such as deeper networks or specialized network designs, could further improve performance.
        Investigating such architectures would require substantial additional effort and represents an interesting direction for future research.
        }


        \textcolor{black}{
        Therefore, SurgiATM can be regarded as a cost-effective solution, requiring no trainable parameters and introducing only minimal computational overhead.
        }

        \subsubsection{\textcolor{black}{Different Dark Channel Usage Strategies}}
        \label{Sec_Dis_Ablation_Different_DCU}

        \begin{table}[t]
            \caption{\textcolor{black}{Experimental results of different dark channel usage strategies. \textbf{Bold} indicates the best metric values for each baseline. \underline{Underline} marks the second-best ones. "*" denotes the third-best ones. "ND" means that the denormalized dark channel $\mathcal{D}$ in SurgiATM is replaced with the normalized dark channel $D$, which is a variant of SurgiATM. "NSD" indicates that the denormalized dark channel in SurgiATM is not smoothed (i.e., $\eta = 0$). "Res" denotes dark-channel-weighted residual learning. "Res" denotes conventional residual learning without using the dark channel. "DPP" indicates that DCP is applied as a post-processing step to refine the predictions of the desmoking network (i.e., the results of "Null"). "DCC" denotes that the dark channel is concatenated with the input image as an additional feature channel. The detailed equations for these usages are summarized in \Cref{Tab_ablation_DC_Eqs}. Since some values are very close to one another, this table reports results to four decimal places.}}
            \label{Tab_ablation_DC}
            \centering
            \resizebox{0.5\textwidth}{!}{
            \tcbox[colframe=red, colback=white, boxrule=0.25mm, size=tight]{\begin{tabular}{l c c c c c}
                \toprule
                Method & DC Usage & CIEDE2000 $\downarrow$ & PSNR $\uparrow$ & RMSE $\downarrow$ & SSIM $\uparrow$ \\ 
                \midrule
                \multirow{7}{*}{AODNet \citep{Li_2017_AOD_Net}} & Null & 10.3495 & 17.3776 & 0.1475 & 0.6930 \\ 
                ~ & Ours & \textbf{6.3270} & \textbf{21.9940} & \textbf{0.0900} & 0.7890* \\ 
                ~ & ND & \underline{6.5670} & 21.4628* & \underline{0.0922} & \underline{0.7925} \\ 
                ~ & NSD & 6.7371* & 21.3831 & 0.0930* & \textbf{0.7960} \\
                ~ & DCR & 7.2047 & \underline{21.7928} & 0.0942 & 0.7910 \\
                ~ & Res & 8.2289 & 20.4390 & 0.1071 & 0.7774 \\ 
                ~ & DPP & 9.7982 & 17.7618 & 0.1351 & 0.6883 \\ 
                ~ & DCC & 9.1528 & 19.1421 & 0.1142 & 0.5755 \\ 
                \midrule
                \multirow{7}{*}{RSTN \citep{Wang_2023_Surgical_smoke}} & Null & 5.3848 & 23.2633 & 0.0736 & 0.7504 \\ 
                ~ & Ours & \textbf{4.9760} & \textbf{24.3060} & \textbf{0.0710} & \underline{0.8240} \\ 
                ~ & ND & \underline{4.9837} & \underline{24.3018} & \underline{0.0711} & \textbf{0.8280} \\ 
                ~ & NSD & 5.0395* & 24.1156* & 0.0717 & 0.8208* \\
                ~ & DCR & 5.1183 & 23.9633 & 0.0715* & 0.8172 \\ 
                ~ & Res & 5.0725 & 24.0167 & 0.0728 & 0.8168 \\ 
                ~ & DPP & 7.1563 & 20.5848 & 0.0994 & 0.7232 \\ 
                ~ & DCC & 5.8639 & 22.8900 & 0.0768 & 0.7260 \\ 
                \midrule
                \multirow{7}{*}{SSIM-PAN \citep{Sidorov_2020_Generative_Smoke}} & Null & 7.0460 & 21.4649 & 0.0898 & 0.6555 \\ 
                ~ & Ours & \textbf{5.8110} & \textbf{23.4770} & \textbf{0.0750} & 0.7500* \\ 
                ~ & ND & 5.8864* & \underline{23.3413} & \underline{0.0759} & \underline{0.7551} \\ 
                ~ & NSD & \underline{5.8149} & 23.2303* & 0.0761* & \textbf{0.7558} \\ 
                ~ & DCR & 6.2599 & 22.9757 & 0.0806 & 0.7322 \\ 
                ~ & Res & 6.3653 & 22.7121 & 0.0807 & 0.7242 \\ 
                ~ & DPP & 8.5076 & 19.6506 & 0.1101 & 0.6271 \\ 
                ~ & DCC & 7.0473 & 21.4602 & 0.0891 & 0.6476 \\ 
                \bottomrule
            \end{tabular}
            }}
        \end{table}

        \begin{table}[t]
            \caption{\textcolor{black}{The equations corresponding to the dark channel (DC) usage strategies evaluated in \Cref{Tab_ablation_DC}. $J_\text{pre}$ denotes the predicted smokeless image. $\Phi(\cdot)$ indicates the original baseline methods.}}
            \label{Tab_ablation_DC_Eqs}
            \centering
            \tcbox[colframe=red, colback=white, boxrule=0.25mm, size=tight]{\begin{tabular}{l c }
                \toprule
                DC Usage & Equation of $J_\text{pre}$ \\ 
                \midrule
                Null & $\Phi(I)$ \\ 
                Ours & $I - \hat{\mathcal{D}} \cdot (1 - \sigma(\Phi(I)))$ \\ 
                ND & $I - D \cdot (1 - \sigma(\Phi(I)))$ \\ 
                DCR & $I + D \cdot \Phi(I)$ \\
                Res & $I + \Phi(I)$ \\ 
                DPP & $\text{DCP}(\Phi(I))$ \\ 
                DCC & $\Phi(I, D)$ \\ 
                \bottomrule
            \end{tabular}
            }
        \end{table}

        \textcolor{black}{
        In terms of different dark channel usage strategies, we summarize the results in \Cref{Tab_ablation_DC}.
        }
        

        \textcolor{black}{
        "DPP" and "DCC" represent straightforward strategies for leveraging the dark channel. "DPP" directly employs DCP as a post-processing step, and "DCC" concatenates the dark channel with RGB images as an additional input channel. These strategies improve the performance of AODNet but degrade the performance of RSTN and SSIM-PAN. This suggests that straightforward dark-channel integration may benefit relatively simple network architectures (e.g., AODNet), but may negatively affect more complex architectures (e.g., RSTN and SSIM-PAN). Therefore, directly applying DCP as a post-processing step or concatenating the dark channel with RGB inputs should be used with caution, particularly for complex network architectures.
        }


        \textcolor{black}{
        Both "Res" and "DCR" introduce superior performance compared with the baseline methods, demonstrating the effectiveness of the residual learning mechanism. For AODNet and SSIM-PAN, "DCR" consistently outperforms "Res", whereas for RSTN, "Res" achieves better performance in terms of CIEDE2000, PSNR, and RMSE. These results suggest that leveraging the dark channel as a weighting factor in residual learning may provide additional benefits for certain restoration networks. Comparing with "Ours", SurgiATM and its variant "ND" consistently achieve superior performance, suggesting the effectiveness of the proposed physics-guided method.
        }


        \textcolor{black}{
        Interestingly, when comparing "Ours" with "ND" and "NSD", the default setting SurgiATM consistently achieves better performance in terms of CIEDE2000, PSNR, and RMSE, whereas its variants "ND" and "NSD" obtain moderately higher SSIM scores. Their overall performances are highly comparable because the only difference is the definition of the used dark channel, while all other settings remain identical.
        To explain the higher SSIM scores achieved by "ND", we attribute this behavior to the estimation of the global illumination $A$ in Eq.~\eqref{Equ_Method_BG_DNN_RES}. Specifically, DCP estimates a global atmospheric light vector for the three color channels by first selecting the pixels with the largest dark-channel values and then choosing the brightest pixel among the corresponding locations in the input image.
        Consequently, the estimated $A$ is shared across the entire image, which may better preserve global structures and thus produce higher structural similarity. 
        However, as discussed in Section~\ref{Sec_Method_DL}, the radial illumination attenuation commonly observed in surgical scenes weakens the assumption that the global illumination $A$ is spatially consistent across all pixels. 
        As a result, this approximation may introduce additional pixel-level errors, leading to inferior performance on pixel-wise metrics such as CIEDE2000, PSNR, and RMSE. 
        In terms of "NSD", the unsmoothed $\mathcal{D}$ can contain zero intensities, thereby leading to gradient vanishing at those pixels. Introducing a small positive offset $\eta$ alleviates this issue by ensuring that all pixels retain a non-zero gradient response. This moderately improves robustness to local- and pixel-level errors, enabling "Ours" to outperform "NSD" on pixel-wise metrics. However, the softened $\mathcal{D}$ may also weaken local contrast and slightly reduce structural consistency, as $\eta$ compresses the dynamic range of $\mathcal{D}$, which may explain the marginal decrease in SSIM.
        }

        \textcolor{black}{
        Therefore, "ND" and "NSD" are both viable alternatives when structural reconstruction accuracy is preferred in SurgiATM. 
        }

        \begin{figure}[t]
        \centering
        \tcbox[colframe=red, colback=white, boxrule=0.25mm, size=tight]{\includegraphics[width=1\linewidth]{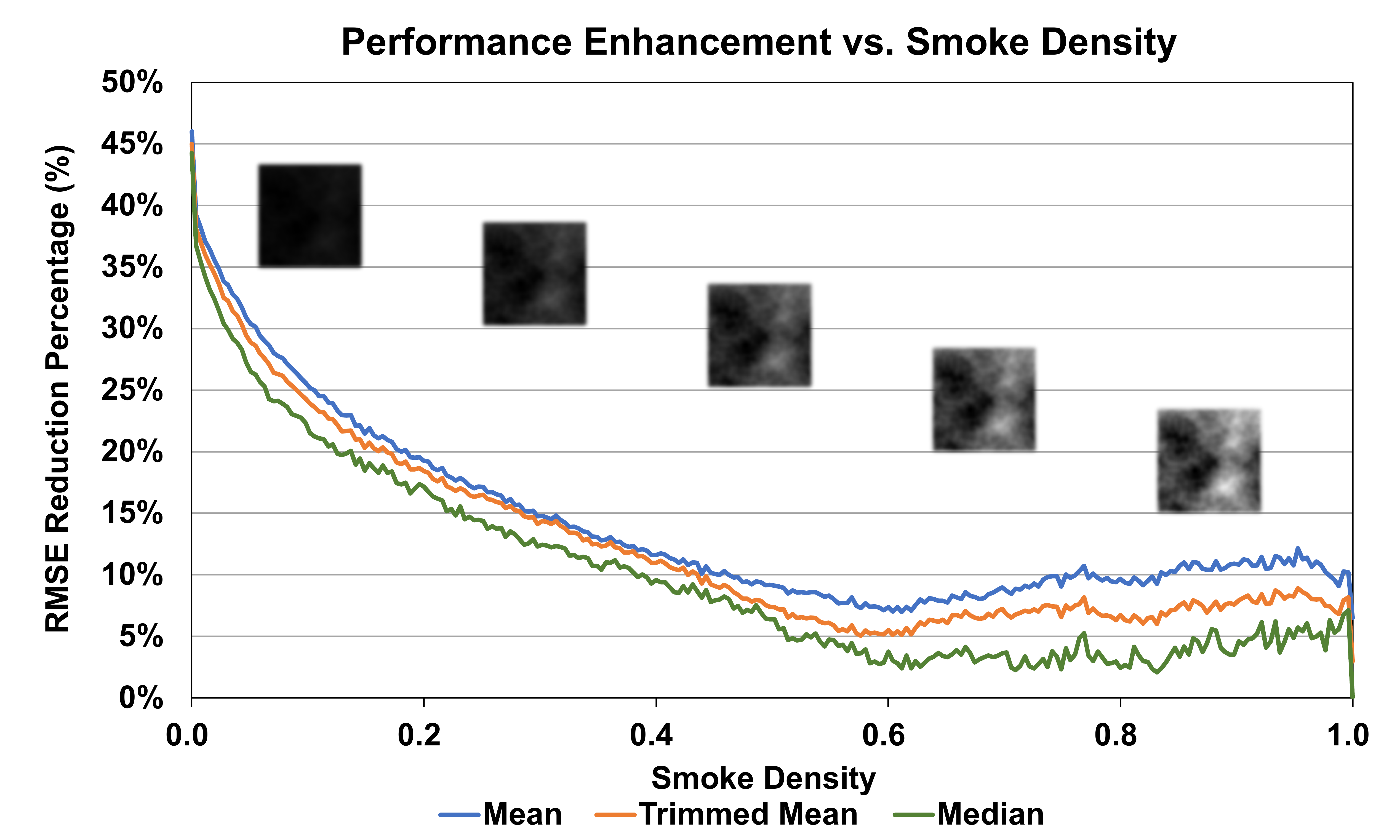}}
        \caption{It shows the average improvement of performance by using SurgiATM in terms of RMSE metrics facing different smoke densities. The data is analyzed pixel-by-pixel from the VASST-desmoke dataset, because it contains ground truth and supports pixel-wise measurement. "Smoke Density" is exactly approximated with the dark channel. The smoke images are generated by Perlin noise to illustrate the smoke density. "Trimmed Mean" computes the mean improvement excluding the max and min values.}
        \label{Fig_Effectiveness_Trend}
        \end{figure}

        \begin{figure}[t]
        \centering
        \tcbox[colframe=red, colback=white, boxrule=0.25mm, size=tight]{\includegraphics[width=1\linewidth]{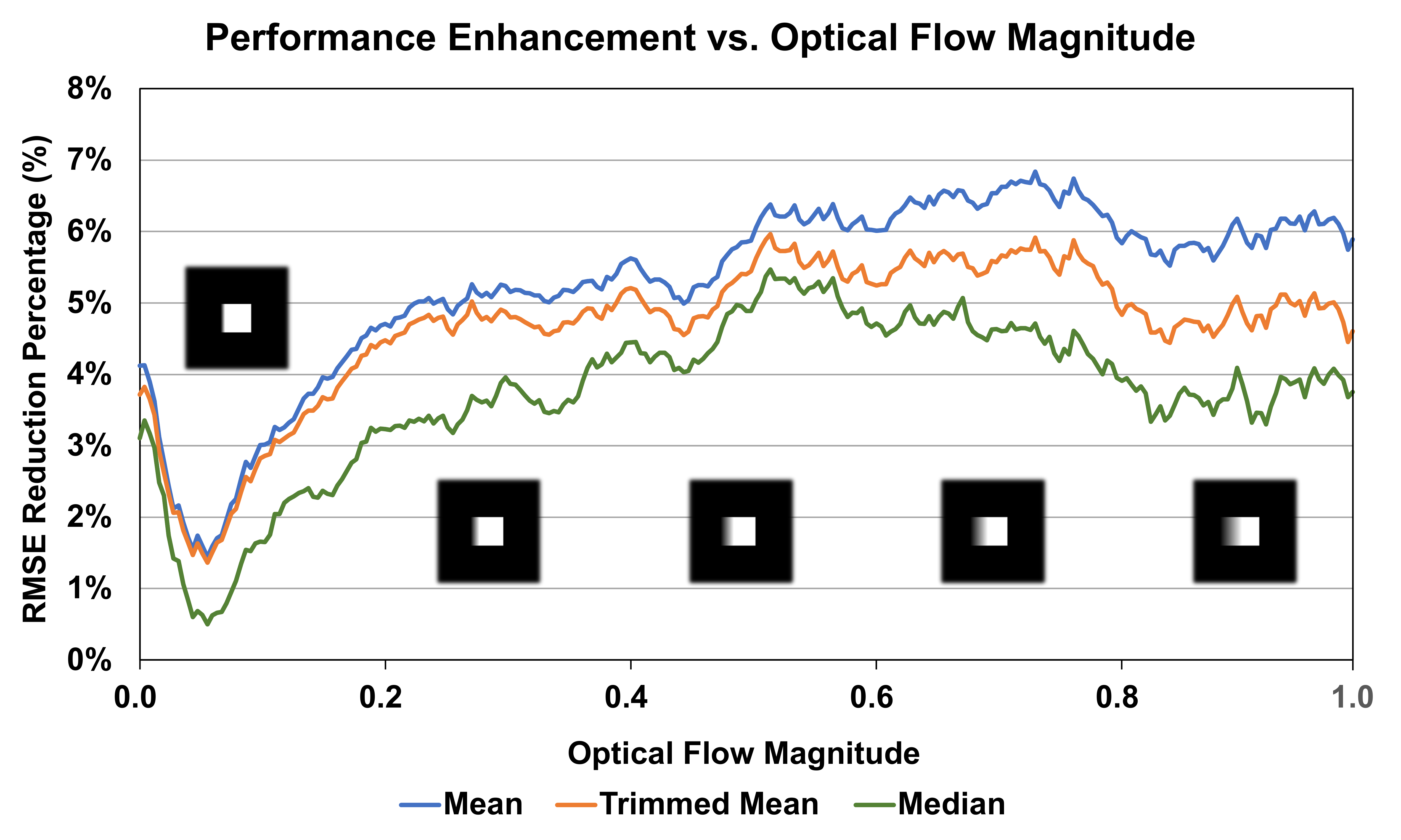}}
        \caption{\textcolor{black}{Performance enhancement of SurgiATM under different optical flow magnitudes. It shows the average performance improvement achieved by SurgiATM in terms of RMSE under different optical flow magnitudes. The results are analyzed on a pixel-by-pixel basis using the Cholec80 dataset (with synthetic smoke), as it contains sufficient frames involving instrument operations and movements. Optical flow is calculated using the Farneback method, and the original magnitude values (empirically ranging from 0 to 20, with values larger than 20 clipped to 20) are normalized to the range of 0 to 1. The square regions are generated through blurring to provide visual references for motion intensity.}}
        \label{Fig_Effectiveness_Trend_Optical}
        \end{figure}

        \subsubsection{\textcolor{black}{Least-Squares-Estimated Gating Weights}}
        \label{Sec_Dis_Ablation_LS_W}

        \textcolor{black}{
        This ablation study is conducted to investigate the performance when the gating weights are estimated using various fitted polynomial functions. We observe that an accurate estimation of the $W^*$ function may lead to superior performance for specific methods and metrics, whereas the approximated $\hat{W}$ provides better overall performance across different settings. Overfitting may be a possible explanation for this performance difference. Results are reported in \ref{app_Dis_Ablation_LS_W}. These results suggest that increasingly accurate approximations of $W^*$ may not yield better restoration for all methods, and that a relatively simple approximation offers a better balance between fitting accuracy and generalization. 
        }

    \subsection{Performance Improvement Analysis}
    \label{Perf_Impr}
        To further investigate the effectiveness of SurgiATM under varying smoke density, we analyze the performance improvements across all baseline methods and show the average enhancement trend in \Cref{Fig_Effectiveness_Trend}. 
        
        
        Overall, SurgiATM enables the baseline methods to generate more accurate frames to varying degrees. Specifically, it provides substantial improvements under thin-smoke conditions and moderate enhancements under dense smoke conditions. An inflection point is observed at a smoke density of approximately 0.5, beyond which the average error reduction stabilizes at around 7.5\%. The performance drop observed at a smoke density of 1.0 is expected and can be attributed to the fact that, in such heavily smoke-contaminated regions, the dark channel value also approaches 1.0. Under this condition, SurgiATM degenerates into ordinary residual learning, making the observed decline theoretically inevitable. The significant improvements under thin-smoke conditions mainly benefit from the ability of consistent prediction that involves minimal or even no color shifting when $\mathcal{D}$ is very small or close to zero, which is more stable and trustworthy compared to learning a pure non-linear mapping with neural networks; this is also a main motivation for using Eq.~\eqref{Equ_Method_RF_WD}.
        
         

        
        \textcolor{black}{
                In addition, we investigate the performance gain with respect to optical flow magnitude (see \Cref{Fig_Effectiveness_Trend_Optical}). Optical flow measures the pixel movement between consecutive frames, allowing rough observation of the movement intensity of objects. 
                The proposed module is not explicitly designed to model motions. Nevertheless, analyzing performance gain across different motion levels provides insight into its robustness under varying scene dynamics. 
                As shown in \Cref{Fig_Effectiveness_Trend_Optical}, SurgiATM slightly enhances the accuracy by around 4\%, consistently achieves improvement over various optical flow magnitudes, and shows a fluctuation without a clear trend, indicating that the effectiveness of the proposed module is not significantly dependent on scene dynamics. 
                Interestingly, optical flow information is not utilized by SurgiATM, while a decrease in performance gain is observed around a normalized optical flow magnitude of approximately 0.1. 
                We hypothesize that moderate motion may introduce ambiguous visual patterns, where subtle motion-induced blur could be partially confused with intrinsic object structures or texture variations.
                Such ambiguity may affect the interaction between the baseline restoration process and the smoke-related prior, resulting in variations in the observed improvement.
        }

    \subsection{Computation Overhead Analysis}
        SurgiATM is a lightweight module, featuring only two hyperparameters and no additional trainable weights. Consequently, its computational cost is minimal, requiring approximately $35$ microseconds ($\mu s$) and $9.8 \times 10^5$ floating-point operations (FLOPs) per frame on our device setup (1 RTX 3090 GPU) with an input resolution of $256 \times 256$. We compute the average runtime across all experiments, and report the incremental overhead introduced by SurgiATM in \Cref{Table_Overhead}. These results demonstrate that SurgiATM has negligible impact on the inference speed of the integrated methods, with overall efficiency primarily determined by the baseline models themselves. This efficiency stems from the fact that SurgiATM is derived from classic physical models and statistical analysis without involving learnable components such as Multilayer Perceptrons (MLPs), thereby introducing minimal parameter overhead.

        \begin{table}[t]
            \caption{Average Incremental Overhead (\%) by Using SurgiATM.}
            \label{Table_Overhead}
            \centering
            \resizebox{0.5\textwidth}{!}{
            \begin{tabular}[width=1.0\linewidth]{l c c }
                \toprule
                Method & $\Delta$ Runtime & $\Delta$ FLOPs \\
                
                \midrule
                
                AODNet \citep{Li_2017_AOD_Net} & 2.0759 & 0.5244 \\
                
                CGAN-DC \citep{Salazar_Colores_2020_Desmoking_Laparoscopy} & 0.6908 & 0.0043 \\
                
                De-smokeGCN \citep{Chen_2020_De_smokeGCN} & 0.2634 & 0.0057 \\
                
                GCANet \citep{Chen_2019_Gated_Context} & 0.3382 & 0.0038 \\
                
                LGUTransformer \citep{Wang_2024_Desmoking_of} & 0.7762 & 0.0011 \\
                
                MARS-GAN \citep{Hong_2023_MARS_GAN} & 0.4748 & 0.0015 \\
                
                MSBDN \citep{Dong_2020_Multi_Scale} & 0.5195 & 0.0025 \\
                
                RSTN \citep{Wang_2023_Surgical_smoke} & 0.4250 & 0.0050 \\
                
                SSIM-PAN \citep{Sidorov_2020_Generative_Smoke} & 0.6661 & 0.0102 \\
                
                SVPNet \citep{Wang_2024_Smoke_veil} & 0.5042 & 0.0011 \\
                
                \midrule

                Avg. & 0.6734 & 0.0560\\
                
                \bottomrule
            \end{tabular}
            }
        \end{table}

    \subsection{Downstream Task Analysis}
    \label{Downstream_Task}
        A high-fidelity surgical scene is a fundamental prerequisite for clinical applications that are largely affected by image quality \citep{Zhao_2025_Rethinking_data}, and smoke removal is expected to support intelligent assistance functions such as surgical scene understanding. To assess the impact of SurgiATM on downstream tasks, we adopt MedSAM \citep{Ma_2024_Segment_anything}, a foundation model for medical scene segmentation, to detect and delineate surgical instruments under different smoke density and light conditions. 
        

        \subsubsection{\textcolor{black}{Downstream Quantitative Results}}

        \begin{table}[t]
        \caption{\textcolor{black}{Downstream surgical instrument segmentation on the CholecSeg8k dataset using MedSAM. The Dice and IoU scores are computed for each frame and then averaged over all frames. "Baseline+" denotes the SurgiATM-integrated version.}}
        \label{Tab_downstream}
        \centering
        \resizebox{0.5\textwidth}{!}{
        \tcbox[colframe=red, colback=white, boxrule=0.25mm, size=tight]{\begin{tabular}{l c c c c}
            \toprule
            \multirow{2}{*}{Method} & \multicolumn{2}{c}{Dice (\%) $\uparrow$} & \multicolumn{2}{c}{IoU (\%) $\uparrow$} \\
            ~ & Baseline & Baseline+ & Baseline & Baseline+ \\ 
            \midrule
            AODNet \citep{Li_2017_AOD_Net} & 89.241 & 90.314 & 82.488 & 83.375 \\ 
            CGAN-DC \citep{Salazar_Colores_2020_Desmoking_Laparoscopy} & 90.935 & 92.016 & 84.798 & 85.728 \\ 
            De-smokeGCN \citep{Chen_2020_De_smokeGCN} & 89.203 & 92.328 & 83.939 & 86.391 \\ 
            GCANet \citep{Chen_2019_Gated_Context} & 93.159 & 93.187 & 87.501 & 87.646 \\ 
            LGUTransformer \citep{Wang_2024_Desmoking_of} & 93.554 & 93.675 & 88.276 & 88.354 \\ 
            MARS-GAN \citep{Hong_2023_MARS_GAN} & 92.693 & 93.208 & 86.924 & 87.586 \\ 
            MSBDN \citep{Dong_2020_Multi_Scale} & 93.372 & 93.441 & 87.372 & 87.981 \\ 
            RSTN \citep{Wang_2023_Surgical_smoke} & 89.726 & 93.174 & 82.936 & 87.551 \\ 
            SSIM-PAN \citep{Sidorov_2020_Generative_Smoke} & 91.580 & 92.717 & 85.711 & 86.843 \\ 
            SVPNet \citep{Wang_2024_Smoke_veil} & 93.741 & 93.985 & 88.468 & 88.856 \\ 
            \bottomrule
        \end{tabular}
        }}
        \end{table}

        \textcolor{black}{
        We leverage CholecSeg8k \citep{Hong_2020_Cholecseg8k_a} and the smoky frame indices provided by \citet{Leibetseder_2017_Real_Time} to sample segmentation data under smoky conditions. The desmoking methods trained on Cholec80 with synthetic smoke are then applied as a pre-processing step before segmentation. The resulting segmentation performances are summarized in \Cref{Tab_downstream}.
        With SurgiATM, the segmentation performance of deep learning-based methods is moderately improved, demonstrating its effectiveness in preserving task-relevant structural information during smoke removal.
        These results suggest that SurgiATM can facilitate downstream surgical scene understanding by alleviating smoke-induced degradation while preserving discriminative visual cues required for segmentation.
        Therefore, downstream evaluation provides complementary evidence to conventional restoration metrics by assessing whether the recovered images remain beneficial for subsequent clinical vision tasks.
        }
        
        \subsubsection{Downstream Qualitative Results}

        As shown in \Cref{Fig_Cholec_to_VASST_MedSAM}, all SOTA methods yield cleaner images and enhance the segmentation performance compared to directly using MedSAM. Moreover, by using SurgiATM, their results are further stabilized and refined to varying degrees, leading to more accurate segmentation on restored images. The results suggest that SurgiATM supports other SOTA desmoking methods to deliver higher overall segmentation accuracy, demonstrating its effectiveness in retaining task-relevant semantic cues that are essential for downstream tasks.

    \subsection{Integration Conflict Analysis}
    \label{Integration_Conflict}

        \subsubsection{Conflict to SVPNet} 
        
        In \Cref{Tab_Cholec_To_VASST}, for SVPNet \citep{Wang_2024_Smoke_veil}, a physics-guided method, SurgiATM causes a slight drop in the SSIM score and minor enhancement in other scores. We investigate the primary mechanism of SVPNet and locate a problem. SVPNet used smoke veil $F \in \mathbb{R}^{W \times H}$ to model the observed frame: 
        \begin{flalign}
            \label{Equ_smoke_veil}
            &\ I(x, c) = F(x) + t(x, c) \cdot J(x, c). &
        \end{flalign}
        Different from other desmoking methods, SVPNet employed three separate U-Nets to estimate $F$, $t$, and $J$ (i.e., multiple outputs), which conflicts with SurgiATM (i.e., single output). When SurgiATM is forcibly integrated with SVPNet (i.e., combining Eqs.~\eqref{Equ_smoke_veil} and \eqref{Equ_Method_RF_RES_3}), the resulting formula becomes unintuitive and difficult to interpret:
        \begin{flalign}
            \label{Equ_conflic_veil_atm}
            &\ I(x, c) = \frac{F(x) - t(x, c) \cdot \mathcal{D}(x) \cdot (1 - \rho(x, c))}{1 - t(x, c)}. &
        \end{flalign}
        Although Eq.~\eqref{Equ_conflic_veil_atm} is differentiable with respect to $F$, $t$, and $\rho$, allowing the neural networks to be updated via backpropagation, the model outputs deviated from their original physical meaning. As a result, it becomes unclear what the networks are actually learning. We hypothesize that this loss of physical interpretability is a main reason why SurgiATM can produce unexpected effects when applied to SVPNet. Consequently, when using SurgiATM, baseline methods with a single output are preferred. Forcibly merging SurgiATM with multi-output approaches may lead to unpredictable or undesired impact (i.e., insignificant improvement or slight decline).

        \subsubsection{\textcolor{black}{Conflict to Diffusion Model}}
        \label{sec_conf_diff}

        \textcolor{black}{
        For fundamental processes of dehazing or desmoking diffusion models such as those proposed by \citet{Chen_2026_EndoIR_Degradation, Chen_2024_LighTDiff_Surgical, Wang_2025_Learning_hazing}, the forward process is:
        \begin{flalign}
        \label{Eq_Diff_FP}
            &\ z_s = \sqrt{\overline{\alpha}_s} \cdot J + \sqrt{1 - \overline{\alpha}_s} \cdot \epsilon, &
        \end{flalign}
        where $J$ is the smokeless image; $z_s$ is the nosied image; $\epsilon \sim \mathcal{N} (0, 1)$; $\overline{\alpha}_s = \prod^s_{i=1} \alpha_i$; and $\alpha_s$ represents the noise scale at each step $s$; and the reverse process is: 
        \begin{flalign}
        \label{Eq_Diff_RP}
            &\ 
            \begin{aligned}
                z_{s - 1} &= \frac{1}{\sqrt{\alpha_s}} \cdot \left( z_s - \frac{1 - \alpha_s}{\sqrt{1 - \overline{\alpha}_s}} \cdot f (z_s, \; I, \; s) \right) \\
                & \quad + \zeta_s \cdot \sqrt{\frac{(1 - \overline{\alpha}_{s - 1}) \cdot (1 - \alpha_s)}{1 - \overline{\alpha}_s}}, 
            \end{aligned} &
        \end{flalign}
        where $f$ is the neural network, $I$ is the smoky image, and $\zeta_s \sim \mathcal{N}(0, 1)$ if $s > 1$, otherwise $\zeta_s = 0$; and the loss function $\mathcal{L}_d$ is:
        \begin{flalign}
        \label{Eq_Diff_LF}
            &\ \mathcal{L}_d = \mathbb{E}_{(J, \, I), \, \epsilon, \, s} \; \left[ \Vert \, \epsilon - f \left(z_s, \; I, \; s \right) \,\Vert^2 \right], &
        \end{flalign}
        On the other hand, the typical loss function for other one-step baseline models (e.g., AODNet, GCANet, etc.) is:
        \begin{flalign}
        \label{Eq_L2_LF}
            &\ \mathcal{L}_2 = \mathbb{E}_{(J, \, I)} \; \left[ \Vert \, J - g \left( I \right) \,\Vert^2 \right] , &
        \end{flalign}
        where $g$ is a neural network, and we use $\mathcal{L}_2$ (i.e., MSE loss) as an example. By comparing Eqs. \eqref{Eq_Diff_LF} and \eqref{Eq_L2_LF}, we notice that diffusion models learn the noise term $\epsilon$ and recover the desmoked image by iteratively removing the estimated noise via Eq. \eqref{Eq_Diff_RP}, whereas non-diffusion models directly predict the clean image $J$. The mappings are represented as: 
        \begin{flalign}
        \label{Eq_DF_MAP}
            &\ f: \mathcal{I} \times \mathcal{Z}_s \times \mathcal{S} \rightarrow \mathcal{E} , &
        \end{flalign}
        and
        \begin{flalign}
        \label{Eq_DE_MAP}
            &\ g: \mathcal{I} \rightarrow \mathcal{J} , &
        \end{flalign}
        where $\mathcal{I}$ and $\mathcal{J}$ represent the smoky and clean image spaces, respectively, $\mathcal{Z}_s$ denotes the noisy image space at diffusion step $s$, $\mathcal{S}$ represents the set of diffusion time steps, and $\mathcal{E}$ denotes the noise space.
        Similarly, the mapping of SurgiATM is: 
        \begin{flalign}
        \label{Eq_ATM_MAP}
            &\ h: \mathcal{I} \times \mathcal{P} \rightarrow \mathcal{J} , &
        \end{flalign}
        where $h$ denotes the proposed method, and $\mathcal{P}$ denotes the output space of the baseline model (i.e., the predicted $\rho$). Its output space naturally aligns with Eq. \eqref{Eq_DE_MAP}, but is inconsistent with Eq. \eqref{Eq_DF_MAP}. 
        In other words, SurgiATM assumes that outputs of the original network lie in the smokeless image space, whereas networks of diffusion models predict noises that are required for the reverse process in the latent diffusion trajectory. 
        Consequently, the current SurgiATM cannot be naturally integrated into $f$ as an output layer due to this fundamental mismatch in mapping. 
        We have not identified an effective solution to this issue, and addressing it remains an important direction for future research. Nevertheless, we perform exploratory experiments to investigate several alternative integration strategies. Although they do not lead to performance improvements, we provide analysis in \ref{app_diffusion_atm}.
        }


\section{Conclusion and Future Work}

    In this work, we propose SurgiATM for laparoscopic surgical smoke removal, which can be seamlessly incorporated into existing DNN-based desmoking methods without additional trainable parameters. \textcolor{black}{By incorporating SurgiATM, these methods achieve performance improvements to varying degrees, while also exhibiting moderately enhanced generalizability. This enables the generation of cleaner frames that are comparatively beneficial for both surgeons and downstream vision tasks.}We analyze the discrepancy between natural dehazing and surgical desmoking and design SurgiATM by first formulating a Mixture-of-Experts model. We then conduct a comprehensive statistical analysis of existing desmoking methods and derive a new restoration formula tailored for surgical smoke removal. Extensive experiments across three distinct datasets validate the effectiveness of SurgiATM. 
    
    
    Nevertheless, several limitations accompany its advantages. The indicator $W$ is currently estimated as a linear function of the dark channel $D$, which captures the general trends across methods, and thus leaves some room for improvement when focusing on a particular method; this could potentially be addressed by deriving method-specific formulations. The denormalized dark channel obtained from DCP remains relatively coarse-grained. Developing an appropriate refinement strategy may further improve its performance and result quality; for example, using guided image filtering or attention mechanisms could generate a comparatively fine-grained denormalized dark channel. \textcolor{black}{
        SurgiATM currently cannot be directly integrated into diffusion-based methods due to the fundamental mismatch between their underlying mappings, and resolving this limitation remains an important avenue for future research. The proposed method mainly addresses smoke degradation, while other complex characteristics of surgical environments, such as occlusions, tissue motion, and confined spaces, are not explicitly considered in the current SurgiATM formulation. Developing a more comprehensive framework that incorporates additional priors to model these factors represents an important direction for future work.
    }


\appendix

\section{\textcolor{black}{Using Gaussian Assumption}}
\label{app_gaussian_w}

\begin{figure}[t]
    \centering
    \tcbox[colframe=red, colback=white, boxrule=0.25mm, size=tight]{\includegraphics[width=1\linewidth]{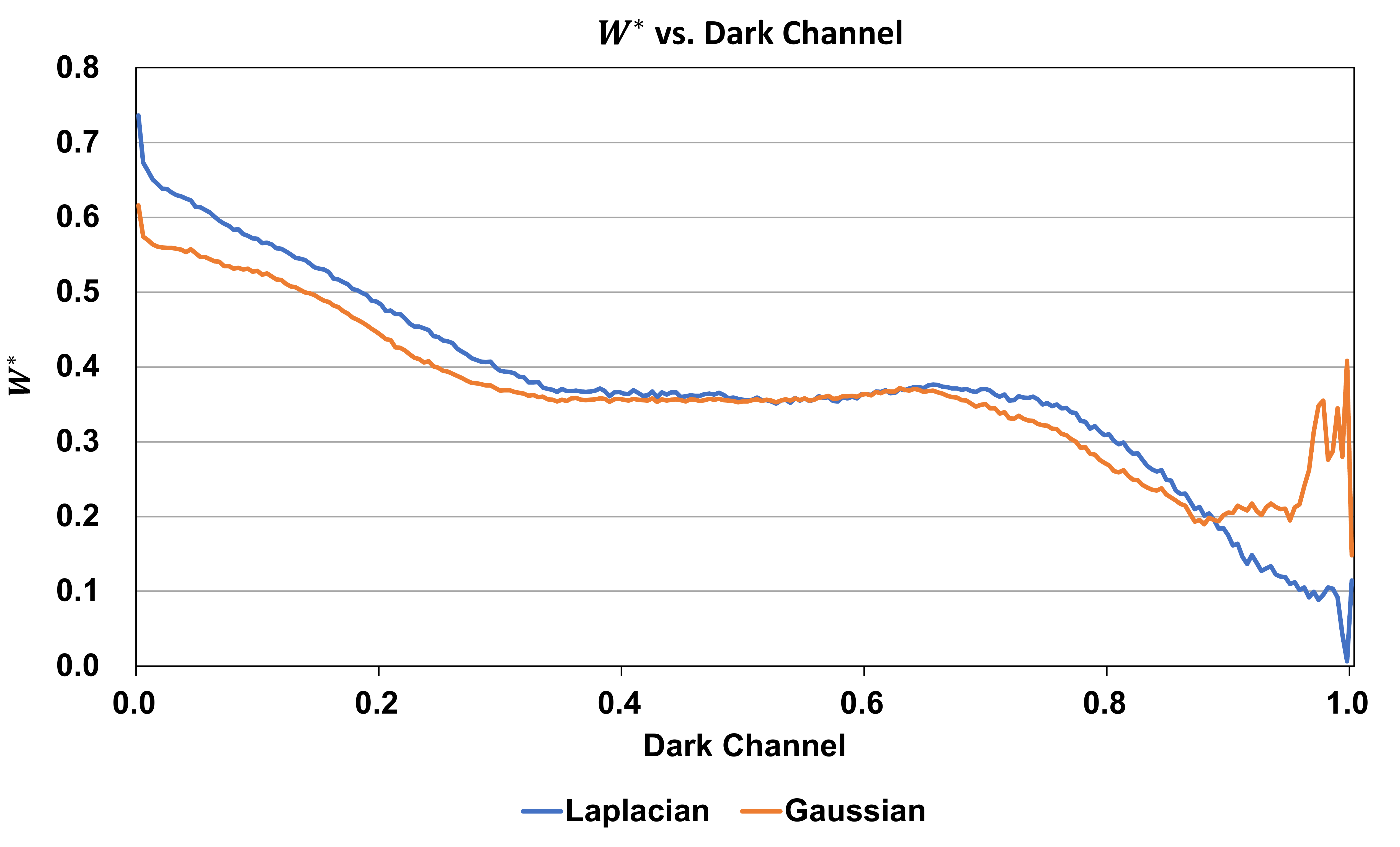}}
    \caption{\textcolor{black}{TEstimated $W^*$ in Gaussian and Laplacian assumptions. The results are averaged from all methods on Cholec80 and VASST-desmoke datasets.}}
    \label{Fig_lap_gau_w}
    \end{figure}

    \textcolor{black}{
    In this section, $W^*$ (see Eq.~\eqref{Equ_Method_EMM_MoE_ERR_OPT_SOLUTION}) is derived from the assumption of a Gaussian distribution instead of a Laplacian distribution. Following the same deduction steps, we have:
    \begin{flalign}
        \label{Equ_Method_EMM_DNN_DCP_ERR_GAU_1}
        &\ \varepsilon_\mathcal{M}(x, c) \overset{\mathrm{i.i.d.}}{\sim} \mathcal{N} \left( \mu_\mathcal{M}(x), \sigma^2_\mathcal{M}(x) \right), &
    \end{flalign}
    and variance and expectation:
    \begin{flalign}
        \label{Equ_Method_EMM_MoE_SIGMA_GAU}
        &\ \begin{aligned}
        \mathrm{Var}[\varepsilon_\mathrm{MoE}] &= W^2(x) \cdot \sigma^2_\mathrm{DCP}(x) \\
            & \quad + (1 - W(x))^2 \cdot \sigma^2_\mathrm{DNN}(x),
        \end{aligned} &
    \end{flalign}
    \begin{flalign}
        \label{Equ_Method_EMM_MoE_MU_GAU}
        &\ \begin{aligned}
        \mathbb{E}[\varepsilon_\mathrm{MoE}] &= W(x) \cdot \mu_\mathrm{DCP}(x) \\
            & \quad + (1 - W(x)) \cdot \mu_\mathrm{DNN}(x).
        \end{aligned} &
    \end{flalign}
    Then, by solving the optimization problem:
    \begin{flalign}
        \label{Equ_Method_EMM_MoE_ERR_OPT_2_GAU}
        &\ \min_{W(x)} \; \; \mathrm{Var}(\varepsilon_\mathrm{MoE}) + \left( \mathbb{E}(\varepsilon_\mathrm{MoE}) \right)^2, & \\
        \label{Equ_Method_EMM_MoE_ERR_OPT_ST1_GAU}
        &\ s.t. \; \; 0 \leq W(x) \leq 1, &
    \end{flalign}
    we obtain the unconstrained solution for the version of the Gaussian assumption:
    \begin{flalign}
        \label{Equ_Method_EMM_MoE_ERR_OPT_SOLUTION_GAU}
        &\ W^*(x) = 
        \frac{\sigma^2_{\mathrm{2}, x} - \mu_{\mathrm{2}, x} \cdot (\mu_{\mathrm{1}, x} - \mu_{\mathrm{2}, x})}
        {\sigma^2_{\mathrm{1}, x} + \sigma^2_{\mathrm{2}, x} + (\mu_{\mathrm{1}, x} - \mu_{\mathrm{2}, x})^2}, &
    \end{flalign}
    where $\sigma_\mathrm{DCP}(x)$, $\sigma_\mathrm{DNN}(x)$, $\mu_\mathrm{DCP}(x)$ and $\mu_\mathrm{DNN}(x)$ are written as $\sigma_{1, x}$, $\sigma_{2, x}$, $\mu_{1, x}$ and $\mu_{2, x}$, respectively. 
    Following Eq.~\eqref{Equ_Method_EMM_MoE_ERR_OPT_SOLUTION_GAU}, the statistical relationship between $W^*$ and $D$ under the Gaussian assumption is illustrated in \Cref{Fig_lap_gau_w}.
    The average Pearson correlation coefficients (PCCs) between $W^*$ and $D$ across all methods and datasets are $-0.870$ and $-0.720$ under the Laplacian and Gaussian assumptions, respectively.
    Consequently, both the Laplacian and Gaussian assumptions exhibit a negative correlation between $W^*$ and $D$, while the Laplacian assumption demonstrates a stronger negative correlation.
    Although the proposed approximation in Eq.~\eqref{Equ_Method_RF_WD} is motivated by the Laplacian assumption, it captures the primary relationship between $W^*$ and $D$ under both statistical assumptions, suggesting that its applicability is not restricted to the Laplacian model alone. This is consistent with the observed compatibility of SurgiATM with a broad range of existing desmoking methods.
    }

\begin{table*}[t]
    \caption{\textcolor{black}{Training and evaluation on VASST-desmoke. \textbf{Bold} represents the best metric score for each diffusion method across different integration strategies. "Null" means the SurgiATM is not applied on the baselines. "SPP" means the SurgiATM is integrated with the baselines via \ref{app_diffusion_atm_c2}. "SOL" indicates the integration in \ref{app_diffusion_atm_c1}.}}
    \label{Table_Diff_Surgi_VASST}
    \centering
    \tcbox[colframe=red, colback=white, boxrule=0.25mm, size=tight]{\begin{tabular}{l c c c c c}
    \toprule
    Method & SurgiATM & CIEDE2000 $\downarrow$ & PSNR $\uparrow$ & RMSE $\downarrow$ & SSIM $\uparrow$ \\
    \midrule
    \multirow{3}{*}{EndoIR \citep{Chen_2026_EndoIR_Degradation}} & Null & \textbf{5.344} & \textbf{23.826} & \textbf{0.072} & \textbf{0.823} \\ 
    ~ & SPP & 5.565 & 23.541 & 0.073 & 0.819 \\ 
    ~ & SOL & 37.162 & 8.343 & 0.383 & -0.010 \\ 
    \midrule
    \multirow{3}{*}{LHTH \citep{Wang_2025_Learning_hazing}} & Null & \textbf{6.765} & \textbf{22.460} & \textbf{0.087} & \textbf{0.785} \\ 
    ~ & SPP & 6.862 & 21.556 & 0.087 & 0.772 \\ 
    ~ & SOL & 37.164 & 8.601 & 0.362 & -0.011 \\ 
    \midrule
    \multirow{3}{*}{LighTDiff \citep{Chen_2024_LighTDiff_Surgical}} & Null & 5.439 & \textbf{23.510} & \textbf{0.073} & \textbf{0.820} \\ 
    ~ & SPP & \textbf{5.376} & 22.768 & 0.077 & 0.759 \\
    ~ & SOL & 36.993 & 8.168 & 0.388 & -0.010 \\ 
    \bottomrule
    \end{tabular}
    }
\end{table*}

\section{\textcolor{black}{Integration with Diffusion Models}}
\label{app_diffusion_atm}

\textcolor{black}{
In this section, we investigate the integration of SurgiATM with diffusion models. Although these integration strategies result in either no performance gains or even performance degradation, they nevertheless offer valuable insights into the underlying conflict between SurgiATM and diffusion models. Experimental results are reported in \Cref{Table_Diff_Surgi_VASST}. The experiment implementation details are the same as those described in Section~\ref{Exp_implement_detail}. Due to device constraints, the batch size for diffusion models is set to 4, instead of 8 or 16 in their original configurations.
}

    \subsection{\textcolor{black}{Using SurgiATM as an Output Layer}}
    \label{app_diffusion_atm_c1}

    \textcolor{black}{
    This integration strategy uses SurgiATM as the output layer of the neural networks in diffusion models. It constitutes an inferior design choice that disregards the mapping inconsistency analyzed in Section~\ref{sec_conf_diff}. This integration is denoted with "SOL" for convenience. Accordingly, the loss function is formulated as:
    \begin{flalign}
    \label{Eq_app_Diff_atm_LF}
        &\ \mathcal{L}_\text{SOL} = \mathbb{E}_{(J, \, I), \, \epsilon, \, s} \; \left[ \Vert \, \epsilon - \left( I - \mathcal{D} \cdot \left( 1 - f(\cdot) \right) \right) \,\Vert^2 \right], &
    \end{flalign}
    where $f(\cdot)$ represents $f \left( z_s, \; I, \; s \right)$ for simplicity. Equation~\eqref{Eq_app_Diff_atm_LF} is differentiable but unintuitive, where SurgiATM is presumed to reconstruct a clean image, but the loss function mandates it to fit noise. Consequently, the reverse process (see Eq.~\ref{Eq_Diff_RP}) is influenced and corrupted into:
    \begin{flalign}
    \label{Eq_Diff_atm_RP}
        &\ 
        \begin{aligned}
            z_{s - 1} &= \frac{1}{\sqrt{\alpha_s}} \cdot \left( z_s - \frac{1 - \alpha_s}{\sqrt{1 - \overline{\alpha}_s}} \cdot \left( I - \mathcal{D} \cdot \left( 1 - f(\cdot) \right) \right) \right) \\
            & \quad + \zeta_s \cdot \sqrt{\frac{(1 - \overline{\alpha}_{s - 1}) \cdot (1 - \alpha_s)}{1 - \overline{\alpha}_s}}, 
        \end{aligned} &
    \end{flalign}
    which is inconsistent with the original reverse diffusion process. For example, when $\mathcal{D} \rightarrow 0$, the process does not remove predicted noise but instead removes an observed image $I$, transforming the "denoising diffusion" into a "de-imaging diffusion", which results in an ill-defined or unpredictable output.
    }


    \textcolor{black}{
    As shown in \Cref{Table_Diff_Surgi_VASST} (see "SOL" rows), the results are consistently inferior to those of the original baseline (i.e., the "Null" rows), confirming that the proposed method cannot be directly appended to the end of the neural networks in diffusion models. SSIM measures the structural similarity between prediction and ground truth, and the negative SSIM scores indicate that the prediction is not similar to the ground truth. This is because the "de-imaging diffusion" negatively influences the diffusion baselines; in other words, the image is gradually erased and thus produces meaningless results.
    }

    \subsection{\textcolor{black}{Using SurgiATM as a Post-Processing Step}}
    \label{app_diffusion_atm_c2}

    \textcolor{black}{
    To avoid the corruption in Eq.~\eqref{Eq_Diff_atm_RP}, SurgiATM can be used as a post-processing step outside of network training, assuming the denoising diffusion process is conducted on normalized radiance $\rho$. This integration is marked with "SPP" for convenience. In this case, Eq.~\eqref{Eq_Diff_FP} is rewritten as:
    \begin{flalign}
    \label{Eq_Diff_atm_FP_c2}
        &\ z_s = \sqrt{\overline{\alpha}_s} \cdot \rho + \sqrt{1 - \overline{\alpha}_s} \cdot \epsilon. &
    \end{flalign}
    Equation~\eqref{Eq_Diff_RP} obtains denoised normalized radiance $\rho_\text{pre} = z_0$, and then use SurgiATM to reconstruct the smokeless image:
    \begin{flalign}
    \label{Eq_Diff_atm_surgiatm_c2}
        &\  J_\text{pre} = I - \hat{\mathcal{D}} \cdot (1 - \rho_\text{pre}), &
    \end{flalign}
    where "pre" indicates the variables are predictions.
    Correspondingly, the target $\rho$ is calculated with reversed SurgiATM:
    \begin{flalign}
    \label{Eq_Diff_atm_rho_c2}
        &\ \rho = 1 - \frac{I - J}{\hat{\mathcal{D}}}. &
    \end{flalign}
    Notably, in Eq.~\eqref{Eq_Diff_atm_rho_c2}, $\rho$ does not represent the realistic normalized radiance. This is because the SurgiATM leverages the normalized radiance to estimate a smokeless prediction, whose reverse process would inevitably involve errors in $\rho$. Consequently, we still name $\rho$ as normalized radiance for convenience, but actually $\rho$ should be interpreted as a latent representation in Eqs.~\eqref{Eq_Diff_atm_FP_c2} and \eqref{Eq_Diff_atm_rho_c2}.
    This integration degenerates SurgiATM into a pre-processing step for the dataloader and a post-processing step for the recovered output, instead of a module component cooperating with neural networks, and therefore has minimal effect on the training trajectory. Furthermore, in Eq.~\eqref{Eq_Diff_atm_rho_c2}, using $\hat{\mathcal{D}}$ as the denominator may lead to numerical instability, thereby adversely affecting the learning quality.
    }


    \textcolor{black}{
    In \Cref{Table_Diff_Surgi_VASST} (see "SPP" rows), compared to the original baseline, incorporating SurgiATM as pre- and post-processing steps around the diffusion process barely produces performance enhancement, demonstrating that this integration strategy offers very little benefit for diffusion models, and even leads to a slight drop in performance.
    }

\section{\textcolor{black}{Challenging Cases}}
\label{app_adverse_cases}

\begin{figure*}[p]
        \centering
        \tcbox[colframe=red, colback=white, boxrule=0.25mm, size=tight]{\includegraphics[width=1.0\linewidth]{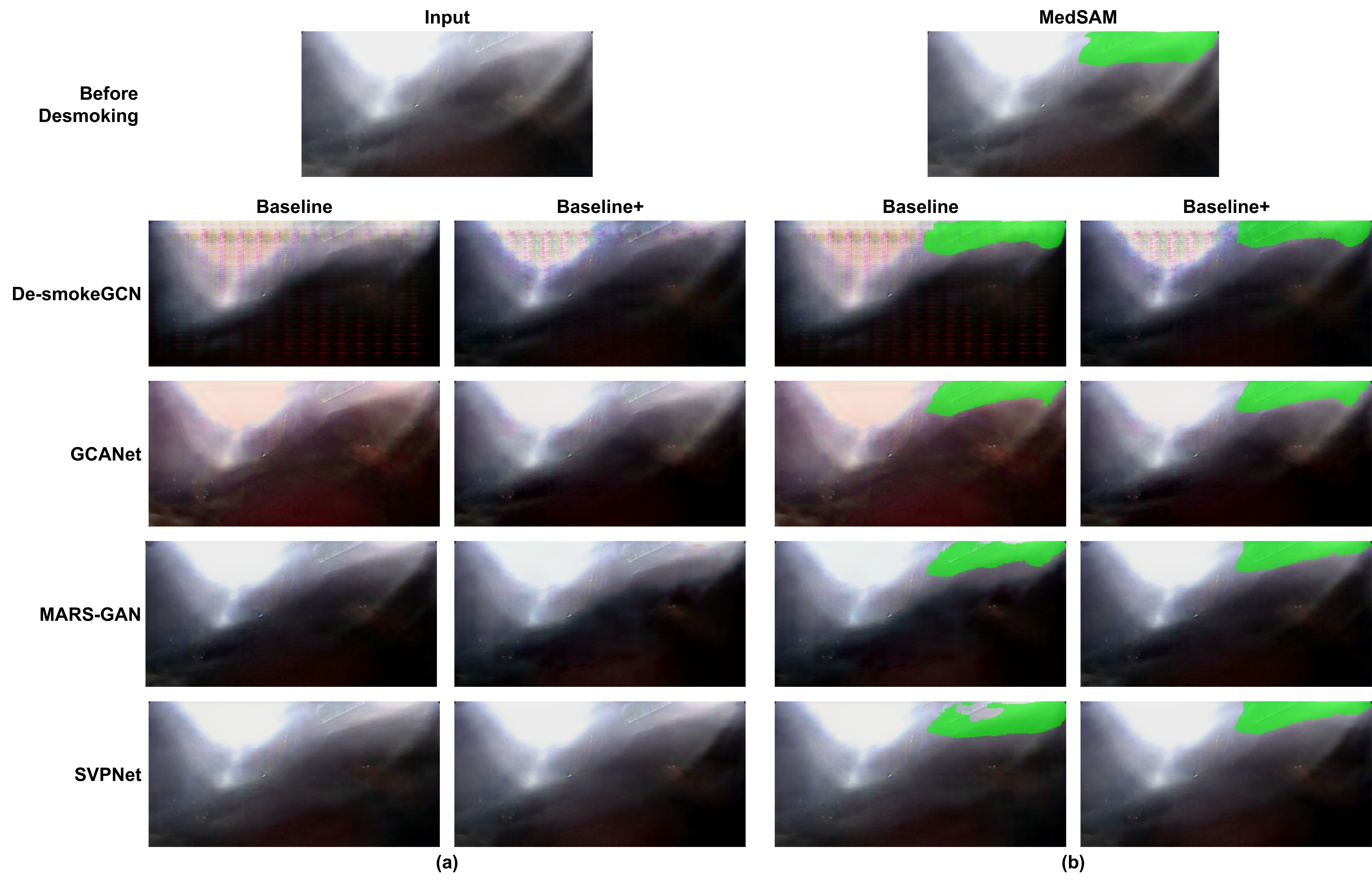}}
        \caption{\textcolor{black}{Heavily smoky frames from the Cholec80 dataset. The top row shows the smoky image and its segmentation before desmoking. (a) shows the desmoking results. (b) presents the segmentation results for each method. "Baseline+" means the SurgiATM-integrated version.}}
        \label{Fig_Adverse_Case}
    \end{figure*}

\textcolor{black}{
To further investigate the behavior of SurgiATM under more challenging conditions, we present additional qualitative comparisons in \Cref{Fig_Adverse_Case}, where the input images contain substantially dense surgical smoke and light reflection. 
Under such challenging conditions, all methods exhibit noticeable degradation, reflecting the inherent difficulty of the task. 
As SurgiATM is designed as a lightweight, trainable-parameter-free module that complements existing desmoking networks rather than replacing them, it provides slight refinements to the baseline predictions instead of fundamentally altering the restoration quality. 
Consequently, the visual differences remain relatively subtle in these adverse cases. 
To better exhibit the benefit of SurgiATM, we leverage MedSAM to generate segmentation on the desmoked frames (see \Cref{Fig_Adverse_Case}(b)), where SurgiATM-integrated baselines contribute to relatively refined masks to different degrees.
}

\begin{table}[t]
    \caption{\textcolor{black}{The equations of least-squares-estimated $W^*$. "PolyN" denotes the number of terms in a polynomial.}}
    \label{Tab_LS_w_eqs}
    \centering
    \resizebox{0.5\textwidth}{!}{
    \tcbox[colframe=red, colback=white, boxrule=0.25mm, size=tight]{\begin{tabular}{l c c }
        \toprule
        Assumption & PolyN & Equation of $W^*$ \\ 
        \midrule
        \multirow{3}{*}{Laplacian} & 1 & $-0.4286 \, D + 0.5891$ \\ 
        ~ & 2 & $0.0135 \, D^2 - 0.4421 \, D + 0.5913$ \\ 
        ~ & 3 & $-2.5437 \, D^3 + 3.8290 \, D^2 - 1.9653 \, D + 0.7170$ \\ 
        \midrule
        \multirow{3}{*}{Gaussian} & 1 & $-0.3022 \, D + 0.5172$ \\ 
        ~ & 2 & $0.1986 \, D^2 - 0.5009 \, D + 0.5502$ \\ 
        ~ & 3 & $-0.9660 \, D^3 + 1.6476 \, D^2 - 1.0793 \, D + 0.5979$ \\ 
        \bottomrule
    \end{tabular}
    }}
\end{table}

\begin{table}[t]
    \caption{\textcolor{black}{Experimental results of different gating strategies. \textbf{Bold} indicates the best metric values for each baseline. \underline{Underline} marks the second-best ones. "*" indicates the third-best ones. "Null" denotes the original baseline. "Ours" represents the baseline augmented with SurgiATM. "Gau1" indicates that the gating weights are estimated by Least Squares using a one-term polynomial under the Gaussian assumption, which also applies to other variants such as "Lap2" and "Lap3". Since some values are very close to one another, this table reports results to four decimal places.}}
    \label{Tab_ablation_LS_W}
    \centering
    \resizebox{0.5\textwidth}{!}{
    \tcbox[colframe=red, colback=white, boxrule=0.25mm, size=tight]{\begin{tabular}{cccccc}
        \toprule
        Method & Gating & CIEDE2000 & PSNR & RMSE & SSIM \\
        \midrule
        \multirow{10}{*}{AODNet \citep{Li_2017_AOD_Net}} & Null & 10.3495 & 17.3776 & 0.1475 & 0.6930 \\ 
        ~ & Ours & \textbf{6.3270} & \textbf{21.9940} & \textbf{0.0900} & \textbf{0.7890} \\ 
        ~ & Gau1 & 7.7725 & 20.7028 & 0.0998 & 0.7484 \\ 
        ~ & Gau2 & 7.7347 & 20.7312 & 0.0998 & 0.7505 \\ 
        ~ & Gau3 & 7.6741 & 20.7816 & 0.0994 & 0.7530 \\ 
        ~ & Lap1 & 7.4766* & 20.9690* & 0.0974* & 0.7606* \\ 
        ~ & Lap2 & 7.4820 & 20.9616 & 0.0975 & 0.7600 \\ 
        ~ & Lap3 & \underline{7.3243} & \underline{21.0788} & \underline{0.0967} & \underline{0.7658} \\ 
        \midrule
        \multirow{10}{*}{RSTN \citep{Wang_2023_Surgical_smoke}} & Null & 5.3848 & 23.2633 & 0.0736 & 0.7504 \\ 
        ~ & Ours & \textbf{4.9760} & \textbf{24.3060} & 0.0710 & \textbf{0.8240} \\ 
        ~ & Gau1 & 5.0953 & 23.9516 & 0.0697 & 0.8093 \\ 
        ~ & Gau2 & 5.1036 & 24.0476 & 0.0695 & 0.8107 \\ 
        ~ & Gau3 & 5.0846 & 24.1183 & \underline{0.0690} & 0.8103 \\ 
        ~ & Lap1 & 5.0825 & 24.1123 & 0.0692 & \underline{0.8127} \\ 
        ~ & Lap2 & 5.0750* & 24.1267* & 0.0691* & 0.8125 \\ 
        ~ & Lap3 & \underline{5.0501} & \underline{24.1783} & \textbf{0.0688} & 0.8126* \\ 
        \midrule
        \multirow{10}{*}{SSIM-PAN \citep{Sidorov_2020_Generative_Smoke}} & Null & 7.0460 & 21.4649 & 0.0898 & 0.6555 \\ 
        ~ & Ours & \textbf{5.8110} & \textbf{23.4770} & \textbf{0.0750} & \textbf{0.7500} \\ 
        ~ & Gau1 & 6.0679 & 22.8895 & 0.0788 & 0.7438 \\ 
        ~ & Gau2 & 6.0914 & 22.6699 & 0.0801 & 0.7455 \\ 
        ~ & Gau3 & 6.0593 & 22.8635* & 0.0799* & 0.7476* \\ 
        ~ & Lap1 & 6.0233* & 22.8517 & 0.0790 & 0.7455 \\ 
        ~ & Lap2 & \underline{5.9708} & \underline{22.8977} & 0.0786 & \underline{0.7492} \\ 
        ~ & Lap3 & 6.0608 & 22.8651 & \underline{0.0783} & 0.7339 \\
        \bottomrule
    \end{tabular}
    }}
\end{table}

\section{\textcolor{black}{Least-Squares-Estimated Gating Weights}}
\label{app_Dis_Ablation_LS_W}

\textcolor{black}{
In this section, SurgiATM is compared with different gating weights fitted by the Least Squares method based on the trends shown in \Cref{Fig_lap_gau_w}. 
The fitted equations of $W^*$ are presented in \Cref{Tab_LS_w_eqs}, and the corresponding experimental results are reported in \Cref{Tab_ablation_LS_W}. In most settings, SurgiATM achieves the best performance, while "Lap3" and "Gau3" exhibit higher RMSE scores for the RSTN baseline. This suggests that the rough approximation of $W$ in Eq.~\ref{Equ_Method_RF_WD} can provide more overall improvements, whereas the relatively accurate estimation obtained by least Squares may achieve better results for specific metrics or specific baselines.
The second-best performances are primarily observed under the Laplacian assumption, suggesting that the Laplacian assumption is relatively more suitable for this work.
}


\section{Visualization Results}
\label{app_is_res}

    \begin{figure*}[p]
        \centering
        \tcbox[colframe=red, colback=white, boxrule=0.25mm, size=tight]{\includegraphics[width=1.0\linewidth]{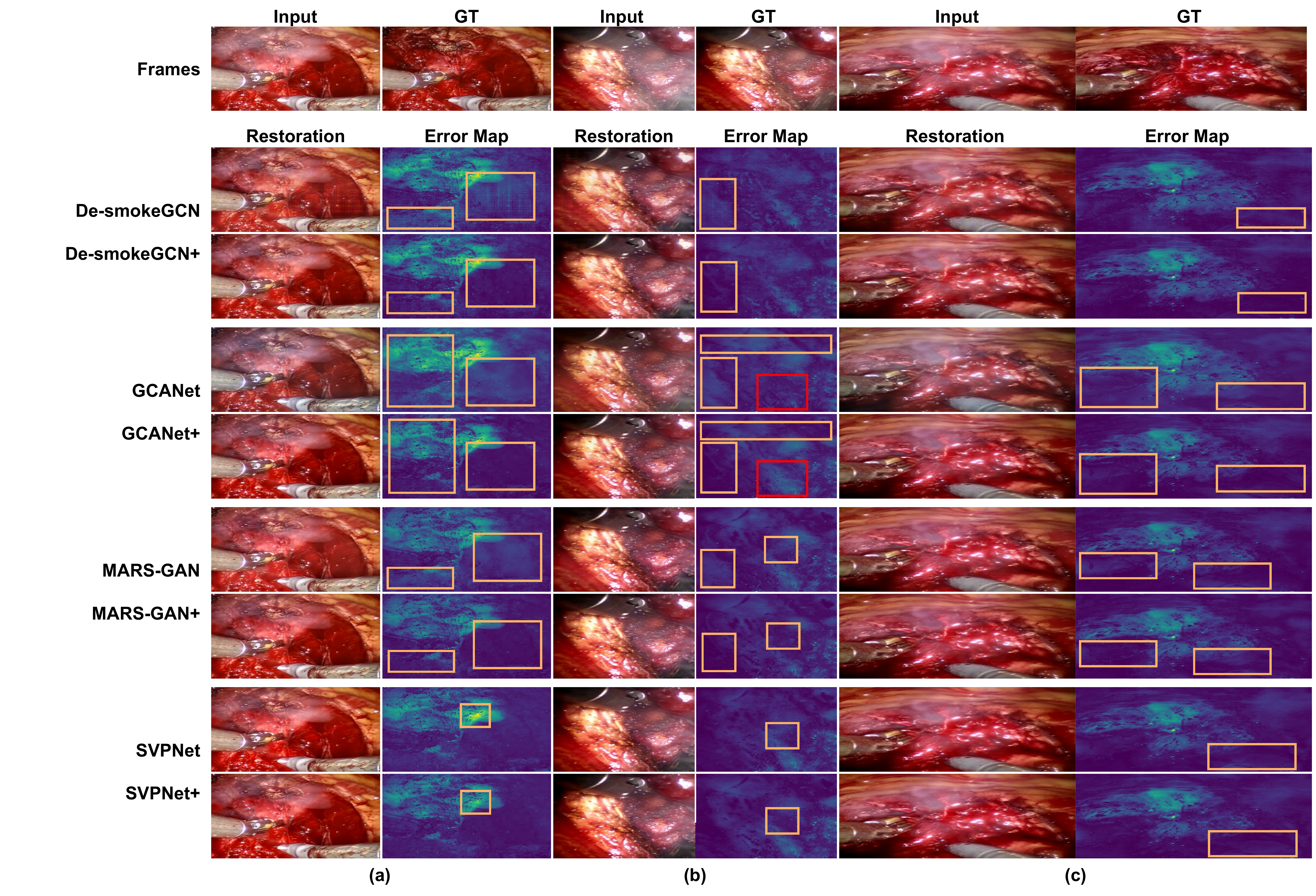}}
        \caption{Error comparison in VASST-desmoke \citep{Xia_2024_A_New}. "+" indicates the baseline integrated with our SurgiATM. "GT" represents ground truth. The methods are trained in Cholec80 \citep{Twinanda_2017_EndoNet_A} with synthetic smoke. Orange boxes mark the errors declined by SurgiATM. Overall, "Baseline + SurgiATM" exhibits more accurate restoration. Red boxes mark the failure cases. "a" shows a frame with adequate light covered by partial dense smoke; "b" displays relatively thin smoke with severe light reflection; "c" represents a frame with both dense smoke and light reflection. The error map demonstrates that SurgiATM can enhance the desmoking results across different conditions.}
        \label{Fig_Cholec_to_VASST}
    \end{figure*}

    \begin{figure*}[p]
    \centering
    \tcbox[colframe=red, colback=white, boxrule=0.25mm, size=tight]{\includegraphics[width=1\linewidth]{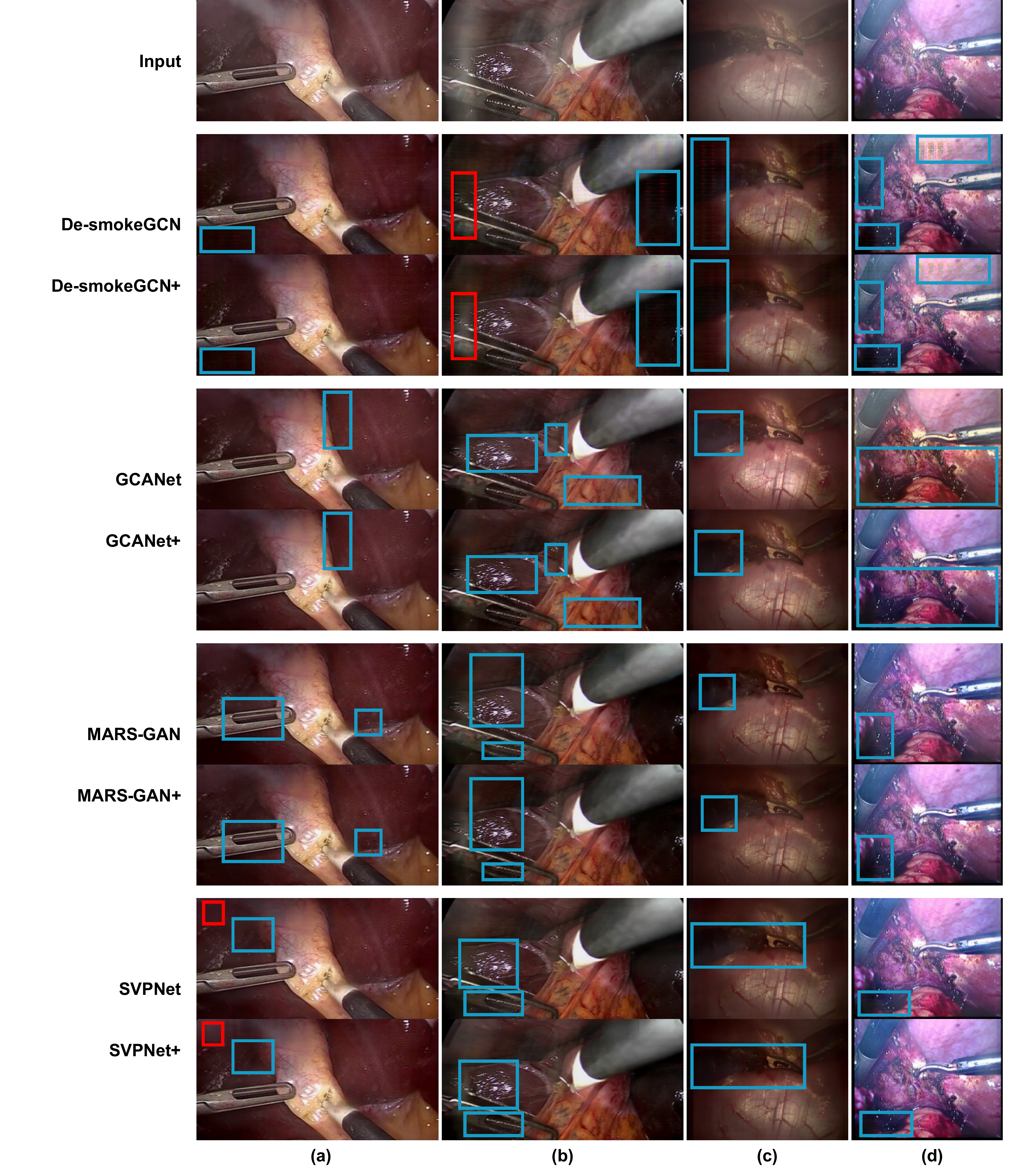}}
    \caption{Comparison of SurgiATM with baselines on real datasets: Cholec80 ("a" and "b") and the Hamlyn Dataset ("c" and "d"). Blue boxes highlight the comparatively accurate and stable restoration achieved by SurgiATM. Red boxes mark the failure cases.}
    \label{Fig_Cholec80_smoky_and_to_Hamlyn}
    \end{figure*}

    \begin{figure*}[p]
    \centering
    \includegraphics[width=1\linewidth]{Cholec_to_VASST_MedSAM.pdf}
    \caption{Qualitative comparisons on the VASST-desmoke dataset with the downstream segmentation task. The top row shows input frames and their segmentation before desmoking. "Baseline" and "Baseline+" represent baseline methods and SurgiATM's version. All segmentation masks are generated by the foundation model MedSAM \citep{Ma_2024_Segment_anything}. (a) shows a comparatively simple scenario with thin smoke, clear instruments, and a plain background; (b) demonstrates a \textcolor{black}{more complex} one where the edge of the instrument is slightly blurred with background; (c) displays a challenging scene with heavy smoke and severe light reflection, where the instrument on the left side is barely visible.}
    \label{Fig_Cholec_to_VASST_MedSAM}
    \end{figure*}

\clearpage
\bibliographystyle{elsarticle-harv} 
\bibliography{main}






\end{document}